%% file: arxiv.tex
\documentclass[10pt,twocolumn,letterpaper]{article}

\usepackage[]{cvpr}      
\input{preamble}
\definecolor{cvprblue}{rgb}{0.21,0.49,0.74}
\usepackage[pagebackref,breaklinks,colorlinks,allcolors=cvprblue]{hyperref}

\let\realcite\cite
\renewcommand{\cite}[1]{\ifx.#1.\hl{[?]}\else\realcite{#1}\fi}
\def\m#1{\mathcal{#1}}

\usepackage{url}
\usepackage{soul,xcolor}
\usepackage{booktabs}
\usepackage{multirow}
\usepackage{colortbl}
\usepackage{enumitem}
\usepackage{graphicx}
\usepackage{mathtools}
\usepackage{subcaption}
\usepackage{tikz}
\newcommand{\model}{\texttt{NORA-1.5}}
\newcommand{\modeltitle}{%
  \textbf{\LARGE%
    \textcolor[HTML]{011B2F}{N}%
    \textcolor[HTML]{011B2F}{O}%
    \textcolor[HTML]{011B2F}{R}%
    \textcolor[HTML]{011B2F}{A}%
    \textcolor[HTML]{011B2F}{-}%
    \textcolor[HTML]{011B2F}{1}%
    \textcolor[HTML]{011B2F}{.}%
    \textcolor[HTML]{011B2F}{5}%
  }%
}
\def\confName{CVPR}
\def\confYear{2026}

\definecolor{cycolor}{RGB}{80,24,134}

\title{\modeltitle{}: A Vision-Language-Action Model Trained using World Model- and Action-based Preference Rewards}

\author{
Chia-Yu Hung$^{1}$ \quad
Navonil Majumder$^{1}$ \quad
Haoyuan Deng$^{1}$ \quad
Liu Renhang$^{1}$ \quad
Yankang Ang$^{1}$ \quad\\
Amir Zadeh$^{2}$ \quad
Chuan Li$^{2}$ \quad
Dorien Herremans$^{3}$ \quad
Ziwei Wang$^{1}$ \quad
Soujanya Poria$^{1}$\\[6pt]
$^{1}$Nanyang Technological University \\
$^{2}$Lambda Labs \\
$^{3}$Singapore University of Technology and Design
}

\begin{document}
\twocolumn[{%
\renewcommand\twocolumn[1][]{#1}%
\maketitle
\includegraphics[width=\textwidth]{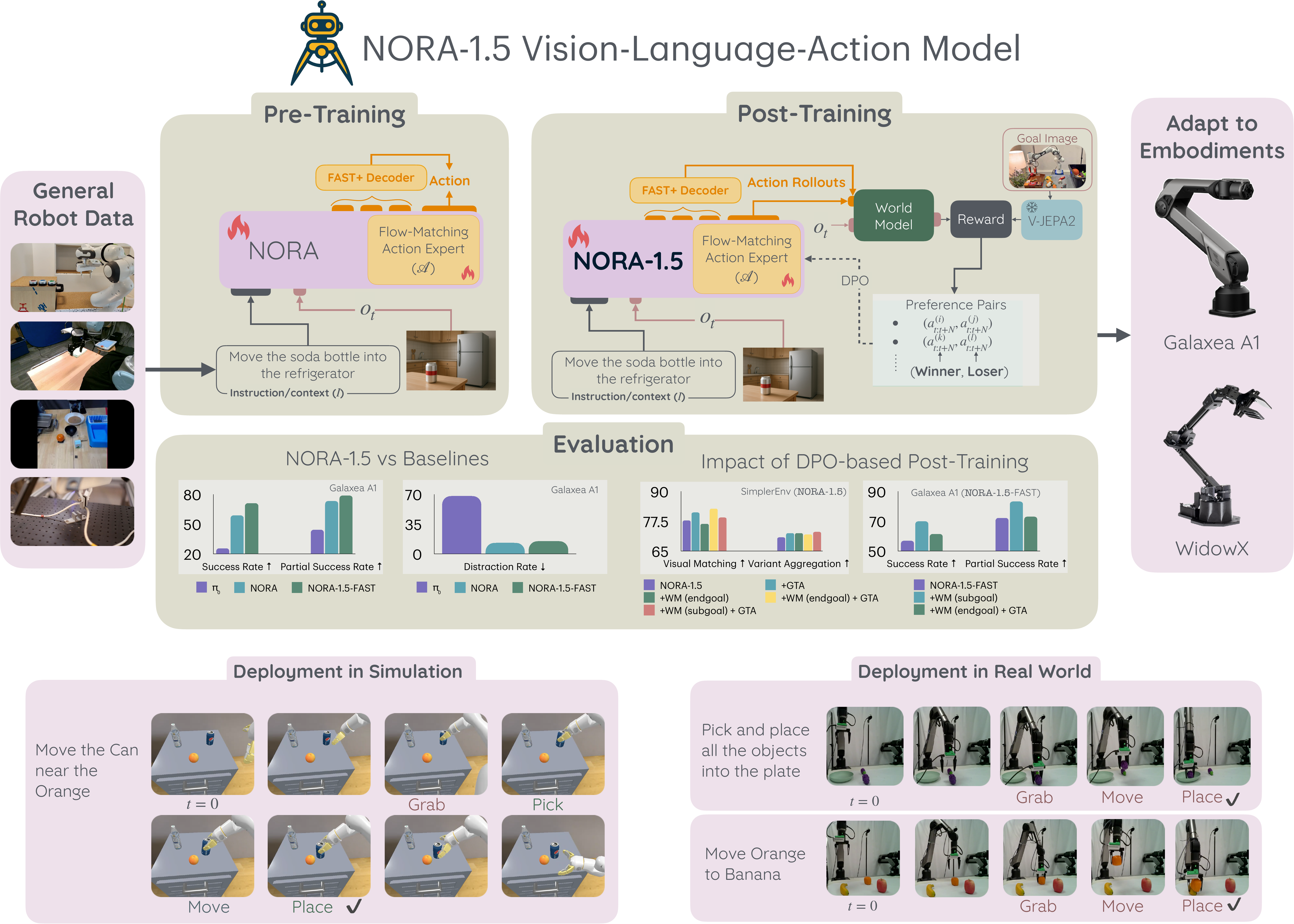}
\label{fig:teaser}
\centerline{\textbf{\textcolor{blue}{\texttt{Project Page:} \url{https://declare-lab.github.io/nora-1.5}}}}
\centerline{\textbf{\textcolor{black}{\texttt{Code:} \url{https://github.com/declare-lab/nora-1.5}}}}
}]

\begin{tikzpicture}[remember picture,overlay,shift={(current page.north west)}]
\node[anchor=north west,xshift=0.4cm,yshift=-2.4cm]{\scalebox{1}[1]{\includegraphics[width=2cm]{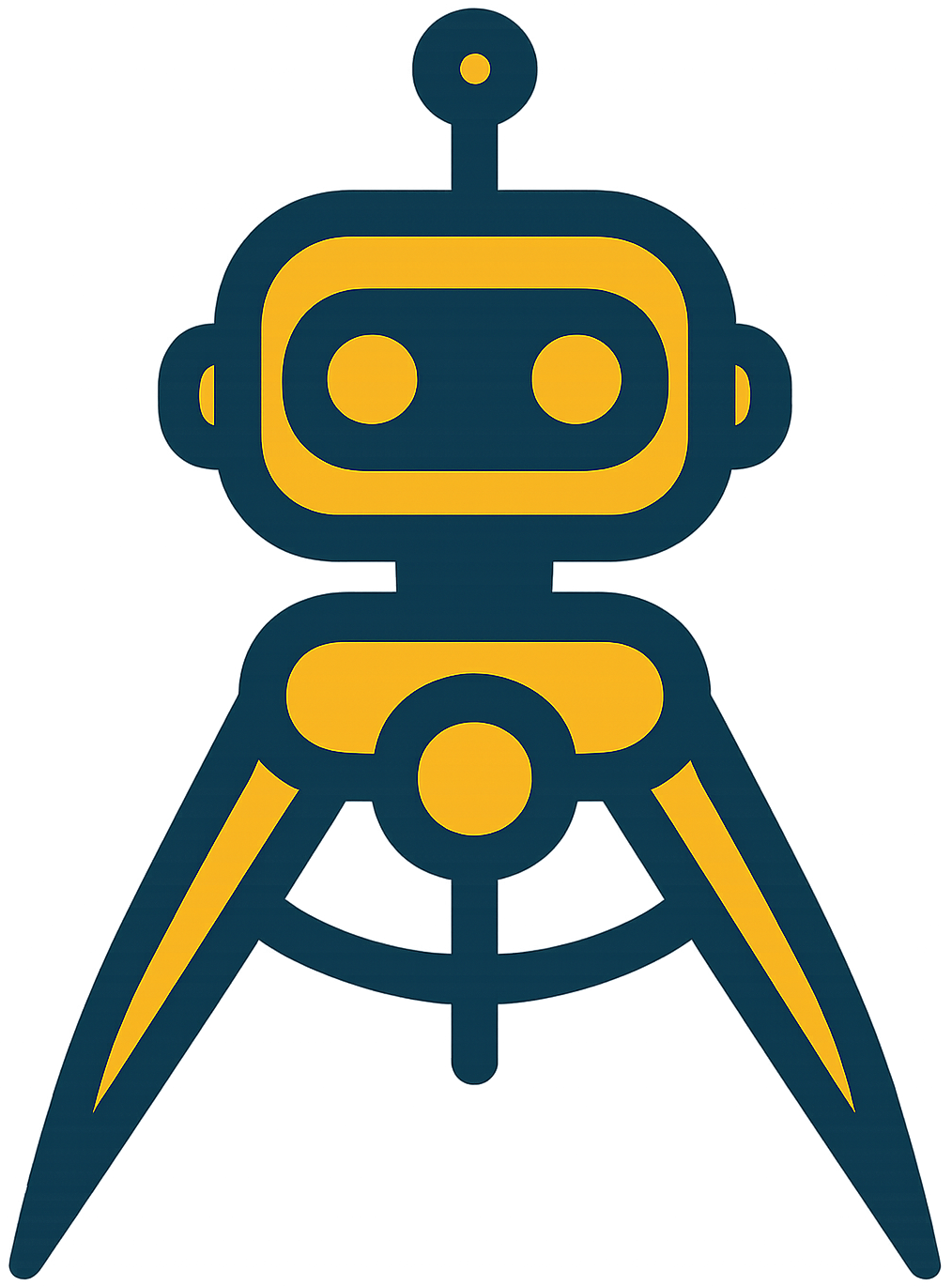}}};
\end{tikzpicture}

\clearpage

\begin{abstract}
Vision--language--action (VLA) models have recently shown promising performance on a variety of embodied tasks, yet they still fall short in reliability and generalization, especially when deployed across different embodiments or real-world environments. In this work, we introduce \model{}, a VLA model built from the pre-trained NORA backbone by adding to it a flow-matching--based action expert. This architectural enhancement alone yields substantial performance gains, enabling \model{} to outperform NORA and several state-of-the-art VLA models across both simulated and real-world benchmarks. To further improve robustness and task success, we develop a set of reward models for post-training VLA policies. Our rewards combine (i) an action-conditioned world model (WM) that evaluates whether generated actions lead toward the desired goal, and (ii) a deviation-from--ground-truth heuristic that distinguishes good actions from poor ones. Using these reward signals, we construct preference datasets and adapt \model{} to target embodiments through direct preference optimization (DPO). Extensive evaluations show that reward-driven post-training consistently improves performance in both simulation and real-robot settings, demonstrating significant VLA model-reliability gains through simple yet effective reward models. Our findings highlight \model{} and reward-guided post-training as a viable path toward more dependable embodied agents suitable for real-world deployment.
\end{abstract}

\section{Introduction}
Recent advancements in Vision--Language--Action Models (VLAs) have demonstrated remarkable performance across a variety of simple embodied tasks, such as picking and placing objects~\cite{pi02024,kim2024openvla,spatialvla2025,sun2025emmax,hung2025nora}. Despite this progress, most existing approaches rely heavily on large-scale imitation learning~\cite{pomerleau1988alvinn} from expert-collected cross-embodiment action trajectories~\cite{open_x_embodiment_rt_x_2023}, followed by supervised fine-tuning (SFT) on embodiment-specific data for downstream tasks. However, SFT-based adaptation inherits a strong bias from limited manually curated demonstrations, restricting the model’s ability to fully generalize or improve beyond the quality of expert data.

To enable more calibrated and scalable post-training, we explore the use of direct preference optimization (DPO)~\cite{rafailov2023direct} by generating preference datasets from reward models capable of ranking the quality of actions produced by the VLA policy.

We introduce \model{}, constructed by coupling a flow-matching--based action expert with the pre-trained autoregressive VLA model NORA~\cite{hung2025nora} through layer-wise self-attention. We choose this architecture due to its promise in achieving impressive performance at a better inference speed as first proposed by \citet{physicalintelligence2025pi05}. While prior work suggested that flow-matching primarily improves inference speed, its impact on policy performance was not investigated. In contrast, we conduct a detailed study and find that flow-matching--based action generation consistently improves performance across multiple benchmarks. We attribute this gain to a strong architectural synergy: the flow-matching expert leverages rich representations encoded by the autoregressive VLA, while the VLA receives informative gradients from the expert, encouraging it to plan coherent multi-step trajectories that the expert can effectively realize. However, we also observe that the flow-matching expert may underperform in low-data settings, likely due to insufficient joint training with the VLA backbone. Overall, \model{} achieves state-of-the-art performance on simulated benchmarks such as SimplerEnv and LIBERO, and its capabilities transfer well to real-world robot experiments on a novel embodiment.

Recent robotics approaches~\cite{levine2016end} have attempted to obtain reward signals by simulating action rollouts. However, such pipelines are computationally expensive, slow to train, and difficult to scale. To address this limitation, we explore reward-driven post-training using lightweight yet effective reward signals derived from compact action-conditioned world models. In this formulation, rewards are estimated by rolling out candidate action sequences through the world model and assessing their ability to reach the goal. Since reward modeling in robotics typically requires estimating how well an action sequence achieves a desired outcome, world models offer a natural mechanism: they directly predict future frames or their latent embeddings conditioned on actions.

Motivated by this, we employ a 1.3B-parameter action-conditioned world model, V-JEPA2-AC~\cite{assran2025vjepa2}, as a goal-based reward estimator. Yet because V-JEPA2-AC is adapted with limited data, its predictions can be noisy. To mitigate this, we incorporate a complementary heuristic reward that measures the distance between sampled actions and ground-truth actions in the training data. These two reward components serve distinct roles: the goal-based world model captures diverse feasible trajectories, while the distance-based heuristic helps counteract noise and provides a stable reference. Across benchmarks, we find that combining these lightweight reward formulations with DPO-based preference tuning consistently improves downstream performance. Once the reward mechanism is in place, it can be applied in multiple ways—including preference optimization (as done here) or reinforcement learning—providing a scalable and data-efficient path for post-training VLAs.

This post-training paradigm defines an economical to scale policy refinement of large Vision–Language–Action (VLA) models. Rather than relying on manually-annotated labels or extensive on-robot rollout execution, \model{} constructs learned evaluators---a world-model based predictor combined with geometric/heuristic checks---that serve as reward proxies to rank model-generated trajectories to form preference pairs; these ranked pairs are then consumed by Direct Preference Optimization (DPO). This approach has three interlinked advantages. Firstly, it converts policy improvement into a compute-bound process: synthetic rollouts sampled from the VLA can be assessed en masse by the learned evaluator, thereby post-training throughput scales with available compute rather than significantly longer physical robot time. Secondly, since DPO optimizes from pairwise preferences rather than exact likelihoods or calibrated densities, the pipeline is naturally compatible with flow-matching or diffusion-based action heads that lack tractable or well-calibrated likelihoods; thus, preference-based objectives avoid a key optimization bottleneck for contemporary VLA architectures.  Thirdly, the learned evaluator provides a unifying rating function that can harmonize heterogeneous corpora: when applied to a corpus such as Open X-Embodiment (OXE), the evaluator may consistently rank trajectories originating from dozens of embodiments, sensors, and task specifications, enabling the entire action-rich structure of OXE to be converted into a massive preference dataset for DPO; the trained world model could be a source of noise this approach. Overall, this enables a single, automated post-training stage that leverages billions of diverse trajectories to produce reward-aligned refinements that generalize across embodiments and deployment conditions---i.e., a scope that meaningfully exceeds mere cross-embodiment adaptation.

In summary, our work makes the following key contributions:
\begin{itemize}[itemsep=0pt, leftmargin=*, wide, labelwidth=0pt, labelindent=0pt, parsep=0pt, topsep=0pt]
    \item \textbf{Introducing \model{}.} We present \model{}, a VLA model built on a strong pre-trained autoregressive VLA (NORA) by integrating a trainable flow-matching action expert and jointly training them on the Open X-Embodiment dataset. \model{} significantly outperforms NORA and achieves state-of-the-art results across diverse simulated benchmarks (SimplerEnv, LIBERO) and real-world embodiments (Galaxea A1).

    \item \textbf{Action-rewarding mechanisms through multiple strategies.} We propose a reward framework composed of (i) goal-based rollouts using an action-conditioned world model (V-JEPA2-AC), (ii) distance-based rewards measuring deviation from ground-truth actions, and (iii) subgoal-based scoring. These complementary signals provide robust and scalable criteria for ranking VLA-generated actions and support DPO-based post-training.

    \item \textbf{Comprehensive architectural analysis.} We conduct a detailed investigation of coupling a flow-matching expert with an autoregressive VLA backbone. Our analysis reveals strong mutual benefits: the expert leverages rich VLA encodings, while the VLA improves its trajectory-level planning through feedback from the expert. We also identify data-regime–dependent behaviors.

    \item \textbf{Advancing scalable post-training of VLAs.} We demonstrate that simple reward models combined with DPO-based preference optimization yield consistent performance gains across both simulation and real robots, establishing a scalable and data-efficient direction for post-training VLA models.
\end{itemize}

\section{Preliminaries}
\label{sec:prelims}

\subsection{NORA}
\label{sec:nora}

NORA~\cite{hung2025norasmallopensourcedgeneralist} is a 3B-parameter auto-regressive Vision-Language-Action (VLA) model obtained by fine-tuning a strong vision language model (VLM) backbone Qwen-2.5-VL-3B~\cite{Qwen2.5-VL} on the Open X-Embodiment dataset~\cite{open_x_embodiment_rt_x_2023} to predict action tokens. Using such a strong VLM backbone imbues NORA with robust world knowledge with multi-modal reasoning, representation learning, and instruction-following capabilities, paramount for natural language- and visuals-driven robotic operations. 
On the other hand, FAST+ tokenizer~\cite{pertsch2025fastefficientactiontokenization} is used for action representation, owing to its efficient discretization of action sequences and proven efficacy~\cite{intelligence2025pi05visionlanguageactionmodelopenworld} across a wide range of action spaces involving single-arm, bi-manual, and mobile robot tasks.

\subsection{V-JEPA-2-AC}

V-JEPA-2-AC~\cite{assran2025vjepa2selfsupervisedvideo} is based on pre-trained V-JEPA-2---a joint embedding architecture model pre-trained by predicting embeddings of the sequence of visual frames from their masked versions. V-JEPA-2-AC uses V-JEPA2 as vision encoder and adds an additional predictor network on top to predict the future frame embeddings, given current frame(s) and a sequence of actions. This model serves as the action-conditioned world model to post-train the VLA models.

\section{\model{}}
\label{sec:nora2}

\begin{figure*}
    \centering
    \includegraphics[width=\textwidth]{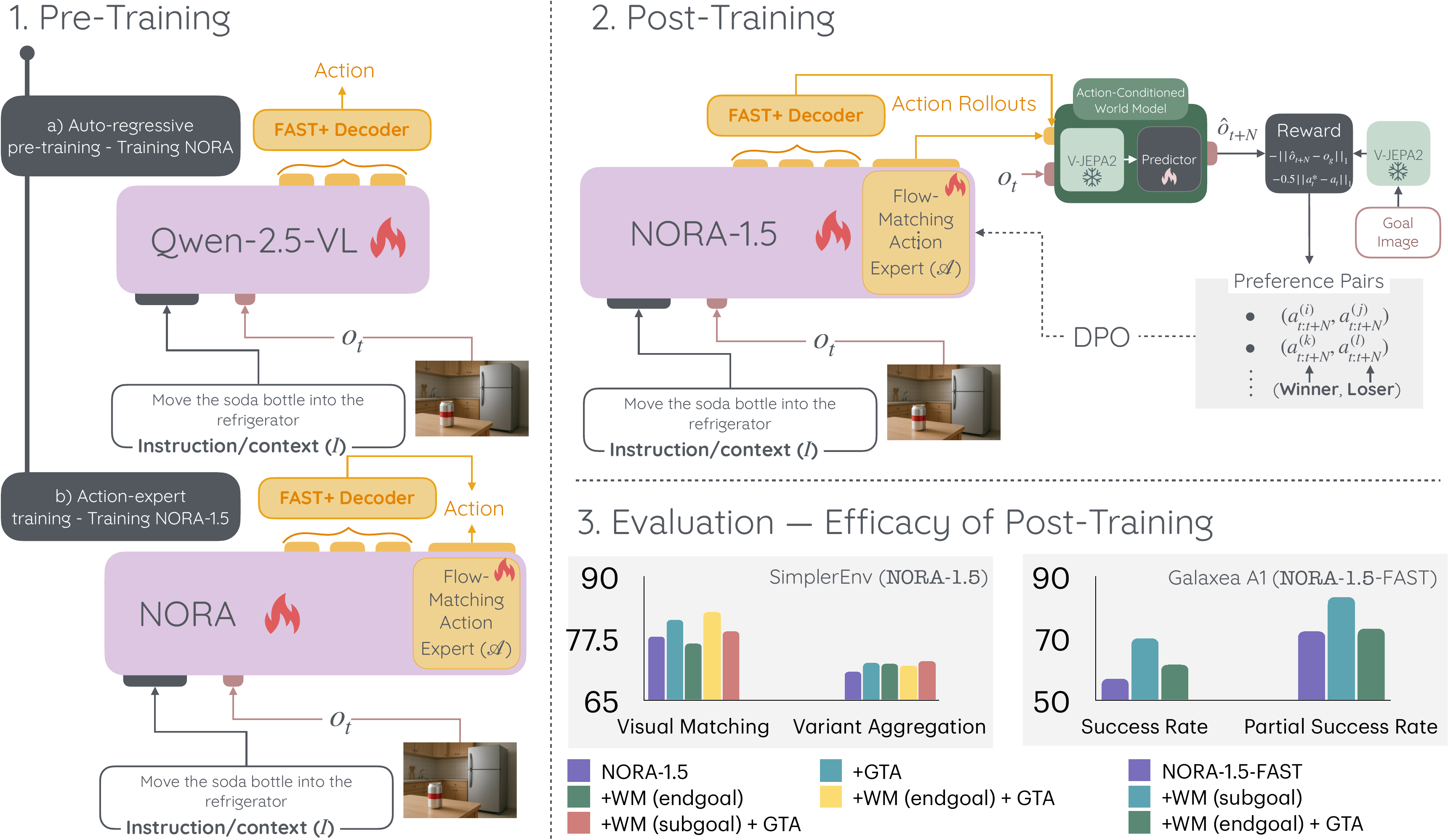}
    \caption{Training pipeline of \model{} where firstly a VLA model is pre-trained through imitation learning and subsequently a preference dataset of the actions is created for preference optimization. \texttt{WM} stands for WM-guided goal-based reward (\cref{eq:goal-re}) and \texttt{GTA} stands for the reward based on ground-truth action (\cref{eq:act-re}).}
    \label{fig:approach}
\end{figure*}

\model{} uses NORA~\cite{hung2025norasmallopensourcedgeneralist} as its VLA/VLM backbone due to its solid vision-language and instruction following capabilities derived from its VLM backbone and imitation learning-based training on a large volume and variety of action trajectories (see \cref{sec:nora}). However, given the performance considerations and efficacy of flow-matching action heads~\cite{intelligence2025pi05visionlanguageactionmodelopenworld}, we add a flow-matching-based dedicated action expert that accepts input from the NORA backbone to generate the action sequences directly. On the other hand, to guide the action generation with a world model---V-JEPA-2~\cite{assran2025vjepa2selfsupervisedvideo} in our experiments---, we align the action expert outputs with direct preference optimization (DPO)~\cite{rafailov2023direct} using the difference between the action-conditioned world model output and the ground-truth frames as a proxy reward. The alignment approach is depicted in \cref{fig:approach}.

\subsection{Architecture}

To circumvent to the often slower auto-regressive action decoding of NORA, we use a separate action expert $\m A$ that directly regresses the action sequence $a_{t:t+N}$ within a horizon of length $N$, based on the joint natural language instruction $I$ and visual observation $o_t$ encoding from NORA ($\m{VL}$):
\begin{flalign}
    K_{\m{VL}, t}, V_{\m{VL}, t} &= \m{VL}_\theta(o_t, I), \label{eq:KV} \\
    a_{t:t+N} &= \m A_\theta(K_{\m{VL}, t}, V_{\m{VL}, t}),
\end{flalign}
where $K_{\m{VL}, t}$ and $V_{\m{VL}, t}$ are the keys and values from the transformer layers of $\m{VL}$. In \model{}, we use the exact same horizon length as NORA-Long, i.e $N= 5$.

\paragraph{Input Encoding with NORA ($\m{VL}$).} Being based on a strong VLM Qwen-2.5-VL-3B allows NORA to have a strong foundation in joint visual-linguistic understanding. Simultaneously, its imitation learning phase on a large volume of diverse trajectories imbued NORA with action generation abilities for a large variety of robots. The latter being an advantage over the typical VLMs makes NORA a good choice for robotics-relevant vision-language encoding to jointly encode natural language instruction and visual observation. To this end, the key-value pairs (see \cref{eq:KV}) of the constituent transformer layers of NORA are used to condition the action expert.

\paragraph{Action Expert ($\m A$).} The action expert is defined as a flow-matching head that regresses the action sequence of horizon $N$, conditioned on key-value pairs of $\m{VL}$. Given the action sequence $a_{t:t+N}$, the noisy action sequence is defined as $a_{t:t+N}^\tau = (1-\tau)a_{t:t+N}+\tau a_0$  where $\tau$ is the flow matching timestep and $a_0 \sim  \mathcal{N}(0,1)$. The action expert $\m A$ directly regresses the ground-truth velocity $v =a_0-a_{t:t+N}$ against the predicted velocity by minimizing the flow matching loss: 
\begin{flalign}
\mathcal{L}_{\text{FM}} = \mathbb{E}_{v,a_{t:t+N}^\tau}  \parallel \m A(a_{t:t+N}^\tau, K_{\m{VL}, t}, V_{\m{VL}, t}) - v \parallel^2  
\label{eq:fm}
\end{flalign}
The vector field regressor $\m A(a_{t:t+N}^\tau, K_{\m{VL}, t}, V_{\m{VL}, t})$ is parameterized as a stacked transformer network, architecturally identical to NORA:
\begin{flalign}
    x^{(l+1)} = Tr^{(l)}(& Q=W_Q^{(l)}x^{(l)}, K=K_{\m{VL}}^{(l)} \oplus W_K^{(l)}x^{(l)},\nonumber\\
    &V=V_{\m{VL}}^{(l)} \oplus W^{(l)}_Vx^{(l)}),
\end{flalign}
where $l$ is the layer index, $x^{(0)} = a_\tau$, $Tr$ is a transformer layer, and $Q, K,$ and $V$ are the query, key, and value inputs to the multi-headed attention therein; the head indices are omitted.  

\subsection{Reward Modeling for Post-training VLAs}

In LLM research, significant gains in System-II level intelligence and task performance have been achieved through extensive post-training using reinforcement learning. The key idea is that the model explores the solution space by generating multiple rollouts. A reward model then evaluates these rollouts based on criteria such as task completion, efficiency, and optimality. The reward signals are used to update the policy, enabling the model to gradually improve its action selection and favor strategies that achieve higher rewards. This process effectively combines exploration of possible actions with guided learning from feedback, allowing the model to discover increasingly effective behaviors. Extending this paradigm to Visual-Language-Action (VLA) models faces a fundamental challenge: how can we define and provide reward signals for these models?  Training a reward model requires data where each action is evaluated based on its affinity to complete the goal successfully.

A naïve strategy would be to sample $N$ action sequences from a VLA model, execute them either in simulation or on a physical robot, and then construct hand-crafted reward signals based on the observed outcomes.  These collected trajectories could then be used to fit reward or value functions capable of evaluating newly generated rollouts and assigning corresponding scores, thereby forming a conventional Reinforcement Learning (RL) pipeline.  In practice, however, this approach presupposes access to highly accurate, fast, and embodiment-specific simulators—or alternatively, to substantial real-robot infrastructure—both of which are costly and often infeasible at scale. As a simpler surrogate, one might instead define rewards by measuring the distance between model-generated actions and their corresponding ground-truth actions; however, such heuristics inherit the limitations of the underlying demonstrations.  In tasks for which multiple valid trajectories exist, distance-based rewards can bias the learner toward a single demonstration path, thereby creating local optima and discouraging exploration of alternative successful behaviors.  Moreover, because these rewards provide no guidance once the policy deviates from the demonstration manifold, they may lead to poor failure recovery and can cause the policy to collapse in off-distribution states encountered during evaluation.

Recent advances in world models and video generative models offer a promising alternative. These models can serve as implicit reward estimators by predicting the consequences of actions and evaluating whether desired subgoals are achieved. Leveraging such learned models as reward functions could enable scalable post-training of VLA policies without the need for fully engineered simulators, providing a practical path forward for reinforcement learning in embodied settings.

\paragraph{Improving the Action Expert through Rewards.}  
Given $N$ rollouts from the action expert, we leverage several techniques to compute rewards as explained in the following section. 

Once the reward model is trained, preference optimization techniques, such as Direct Preference Optimization (DPO)~\cite{rafailov2023direct}, RL, and GRPO (Group Reward Preference Optimization)~\cite{shao2024deepseekmathpushinglimitsmathematical}, can be adopted for improving the action expert. In our case, we use DPO.

\paragraph{Reward Designs.} Our reward model has two components: (i) WM-guided goal-based reward and (ii) action-based reward.
\emph{WM-guided goal-based reward} is designed to quantify the alignment of the generated actions to the specified goal. For this, action-conditioned world models can be used to predict the resulting future states. These states can then be compared to the ground-truth goal states---we experimented with both final goal, denoted as \texttt{WM} (endgoal) reward, and immediate subgoal states, denoted as \texttt{WM} (subgoal) reward (see \cref{sec:ablation})---using a suitable metric to obtain the reward signal for the action expert. The immediate subgoal states could guide the model toward immediate short-term goals, as opposed to the end-goal state that could guide toward the final long-term goal. Following this hypothesis, to estimate the quality of the actions in terms of achieving the end-goal or subgoal, we train an action conditioned world dynamics model $\m W$ that is based on pre-trained V-JEPA2\footnote{\url{https://dl.fbaipublicfiles.com/vjepa2/vitg.pt}}---trained to encode images and sequence of images. Inspired by \citet{assran2025vjepa2selfsupervisedvideo}, we train a predictor transformer model ($P_\theta$) that accepts the current observation $o_t$ encoded by V-JEPA2 ($\m J$) and an action sequence $a_{t:t+N}$ as input, to regress the embedding of the next observation $\hat o_{t+N}$, as defined in \cref{eq:wm}.
\begin{flalign}
    & \m J(o_{t+N}) = \m W_\theta(o_t, a_{t:t+N}) \coloneq \m P_\theta(\m J(o_t), a_{t:t+N}), \label{eq:wm} \\
    & \m R_g(a_{t:t+N}, o_t) \coloneq - ||\m J(o_g) - \m W_\theta(o_t, a_{t:t+N})||_1 \label{eq:goal-re}, \\
    & ~~~~~~~~~~~~~~~~~~~~~~~~~~~~~~g\in \{\text{endgoal}, \text{subgoal-}t\}, \nonumber \\
    & \m R_a(a_{t:t+N}) \coloneq - || a^*_{t:t+N} - a_{t:t+N}||_1 \label{eq:act-re}, \\
    & \m R_\text{tot}(a_{t:t+N}, o_t) \coloneq \m R_g(a_{t:t+N}, o_t) + 0.5 \m R_a(a_{t:t+N}) \label{eq:tot-re}
\end{flalign}
The WM-guided goal-based reward, as defined in \cref{eq:goal-re}, is the difference between the final goal image $o_\text{endgoal}$ or the immediate subgoal image $o_{\text{subgoal-}t}$ and the world model-estimated resultant image of the candidate action $a_{t:t+N}$. This difference could indicate how close an action $a_t$ is to take the task to the end-goal or the immediate subgoal. The ground-truth subgoal image $o_{\text{subgoal-}t}$ at time $t$ is chosen as the $t+N$-th available frame $o_{t+N}$.

On the other hand, \emph{action-based reward}~\cite{kwok2025robomonkey} (referred to as \texttt{GTA} in the experimental results), as defined in \cref{eq:act-re}, quantifies how close an action $a_{t:t+N}$ is to the gold action $a^*_{t:t+N}$. The total reward $\m R_\text{tot}$ combines these two components, where the action-based reward is given half as much weight as the WM-guided goal-based reward. This combination could mitigate the noisiness of the WM-guided goal-based reward inherited from the action-conditioned world model $\m W$ that is trained on limited data and may not generalize well to all scenarios. On the other hand, action-based reward can be too constrained, as the ground-truth trajectory may not be unique, and in such cases, goal-driven reward may work well.

The reward model used in this work provides dense, stepwise evaluations which permit the model to rank sampled candidate actions at each timestep.  Concretely, given a fixed task specification and observation $s_t$, the model assigns comparative scores to different candidate actions $\{a^{(1)}_{t:t+N},\dots,a^{(N)}_{t:t+N}\}$, enabling the VLA to discriminate the relative quality of these actions and thereby encouraging deeper step-level exploration during Direct Preference Optimization (DPO).  Because the ranking is performed at the action level, the policy can explore diverse local decision branches and propagate preference information that is localized in time.  This reward model can also be integrated directly into conventional RL objectives (e.g., as per-step rewards $r_t$ or as an auxiliary critic), enabling hybrid training regimes.  

By contrast, an alternative is to collect data where we use sparse per-step rewards derived from the final trajectory outcome, use that to train a value function, and finally to perform RL through learned value functions; while repeated trajectory-level rollouts also promote exploration, they generally yield shallower exploration because credit is assigned over whole trajectories rather than to individual time steps. 

\paragraph{Preference Dataset Construction.} We construct preference datasets $\m D_\text{goal}$ and $\m D_\text{act}$ of (\texttt{winner}, \texttt{loser}) action preference pairs $(a_{t:t+N}^W, a_{t:t+N}^L)$ based on rewards defined in \cref{eq:act-re,eq:tot-re}, respectively, where $a_{t:t+N}^{W, L} \sim \mathrm{VLA}_\theta(o_t, I)$, $\mathrm{VLA} \coloneq \m A_\theta \circ \m{VL}_\theta $, and $\m R(a_{t:t+N}^W,\cdot) > \m R(a_{t:t+N}^L,\cdot)$. Given the current state, instruction, and these pairs, we use \cref{eq:goal-re}, \cref{eq:act-re}, and \cref{eq:tot-re} to rank the actions given an observation and construct the preference pairs accordingly.

\subsection{Training}

There are two major training stages:
\begin{enumerate}[itemsep=0pt, leftmargin=*, wide, labelwidth=0pt, labelindent=0pt, parsep=0pt, topsep=0pt, label=\roman*.]
    \item \textbf{Action-Expert Training.} The action expert parameters are randomly initialized and subsequently jointly trained with the VLA-backbone (NORA) parameters with a combined flow-matching loss on the action expert output and cross-entropy loss on the FAST+ output tokens of NORA. 
    
    \item \textbf{Reward-guided Post-Training.} We align the action expert-generated action sequences with DPO objective
    \begin{flalign}
        \small
        L_{\text{DPO-FM}} = & -\mathbb{E}_{\tau\sim \mathcal{U}(0, 1), (a^W_{t:t+N}, a^L_{t:t+N}, o_t, I) \sim \m D_.} \nonumber \\
        \log \sigma \Big( &-\beta  \Big[ 
        \underbrace{\|\m A(a^W_{t:t+N}, o_t, I, \tau; \theta) - v^W_\tau\|_2^2}_{\text{Winning loss}} \nonumber \\
        & - \underbrace{\| \m A(a^L_{t:t+N}, o_t, I, \tau; \theta) - v^L_\tau \|_2^2}_{\text{Losing loss}} \nonumber \\
        &-  \underbrace{\|\m A(a^W_{t:t+N}, o_t, I, \tau; \theta_{\text{r}}) - v^W_\tau\|_2^2}_{\text{Winning reference loss}} \nonumber\\
            &+ \underbrace{\|\m A(a^L_{t:t+N}, o_t, I, \tau; \theta_{\text{r}}) - v^L_\tau \|_2^2}_{\text{Losing reference loss}}
    \Big]
\Big).
    \end{flalign}
    On the other hand, we also align the FAST+ action outputs from the VLA decoder head with the DPO objective by \citet{rafailov2023direct}. The evaluations of FAST+ outputs are indicated with a `-FAST' suffix. The DPO-based post-training is applied to the SFT models i.e., after fine-tuning the VLA on target embodiment's supervised data.
\end{enumerate}

\section{Experiments}
\label{sec:exp}
\subsection{Baselines}
As baselines, we use existing well known VLA models including autoregressive VLAs such as SpatialVLA~\cite{spatialvla2025}, RT-1~\cite{rt12022arxiv}, MolmoAct~\cite{lee2025molmoactactionreasoningmodels}, Emma-X~\cite{sun2024emmaxembodiedmultimodalaction}, NORA~\cite{hung2025norasmallopensourcedgeneralist}, and OpenVLA~\cite{kim2024openvla} and diffusion or flow-matching based such as  $\pi_0$~\cite{pi02024} etc. 
\subsection{Benchmarks and Evaluation Settings}
The VLA models are evaluated across both simulated and real-world settings using LIBERO, SimplerEnv, and a Galaxea A1 robotic arm. The LIBERO benchmark comprises four subsets---Spatial, Object, Goal, and Long---each evaluated over 500 episodes, with results averaged across three runs using different seeds; fine-tuning is performed by combining data from all four subsets after removing no-op actions. SimplerEnv focuses on closing the simulation--reality gap with optimized PD parameters and evaluates four tasks (pick coke can, move object near object, open drawer, close drawer) under two protocols: visual matching and variant aggregation, covering over 1{,}000 episodes in total; results are averaged over two runs. For real-world cross-embodiment evaluation, we use the Galaxea A1 robot---absent from the pretraining dataset---and collect 1{,}000 teleoperated pick-and-place episodes with randomized object placement across nine unique tasks (e.g., ``Put apple on the plate''). Evaluation is conducted on nine tasks grouped into three categories (seen tasks, unseen-object--seen-distractor tasks, and unseen-instruction--seen-distractor tasks), each repeated for 10 trials using fixed starting positions consistent across baselines. Simulation benchmarks report binary success rates (1 if the task is completed, else 0), while real-robot evaluations report both success rate (\emph{Succ.}$\uparrow$) and partial success rate (\emph{Part. Succ.}$\uparrow$) to capture finer-grained differences in performance i.e., if the robot successfully grasps the correct object, we reward it with one point. Additionally, we also report the occurrences of grasping the distractors (\textcolor{red}{Dist.}$\downarrow$) in the environment where a lower score is preferred.

\subsection{Performance of \model{}}

\begin{table*}[ht]
\caption{Performance comparison across models on SimplerEnv evaluation. The baseline results are taken from \citet{lee2025molmoactactionreasoningmodels}. VM:=Visual Matching, VA:=Variant Aggregation, PCC:=Pick Coke Can, MN:=Move Near, DR:=Open/Close Drawer.}
\label{tab:simplerenv-eval}
\centering
\resizebox{\linewidth}{!}{
\begin{tabular}{lccc>{\columncolor{gray!10}}cccc>{\columncolor{gray!10}}c}
\toprule
\multirow{2}{*}{\textbf{Model}} & 
\multicolumn{4}{c}{\textbf{Visual Matching}} & 
\multicolumn{4}{c}{\textbf{Variant Aggregation}} \\
\cmidrule(lr){2-5} \cmidrule(lr){6-9}
& Pick Coke Can & Move Near & Open/Close Drawer & Avg 
& Pick Coke Can & Move Near & Open/Close Drawer & Avg \\
\midrule
HPT  & 56.0\% & 60.0\% & 24.0\% & 46.0\% & --- & --- &--- &--- \\
TraceVLA  & 28.0\% & 53.7\% & 57.0\% & 42.0\% & 60.0\% & 56.4\% & 31.0\% & 45.0\% \\
RT-1-X  & 56.7\% & 31.7\% & 59.7\% & 53.4\% & 49.0\% & 32.3\% & 29.4\% & 39.6\% \\
RT-2-X  & 78.7\% & 77.9\% & 25.0\% & 60.7\% & 82.3\% & 79.2\% & 35.3\% & 64.3\% \\
Octo-Base  & 17.0\% & 4.2\% & 22.7\% & 16.8\% & 0.6\% & 3.1\% & 1.1\% & 1.1\% \\
OpenVLA  & 16.3\% & 46.2\% & 35.6\% & 27.7\% & 54.5\% & 47.7\% & 17.7\% & 39.8\% \\
RoboVLM (zero-shot)  & 72.7\% & 66.3\% & 26.8\% & 56.3\% & 68.3\% & 56.0\% & 8.5\% & 46.3\% \\
RoboVLM (fine-tuned) & 77.3\% & 61.7\% & 43.5\% & 63.4\% & 75.6\% & 60.0\% & 10.6\% & 51.3\% \\
Emma-X  & 2.3\% & 3.3\% & 18.3\% & 8.0\% & 5.3\% & 7.3\% & 20.5\% & 11.0\% \\
Magma  & 56.0\% & 65.4\% & \textbf{83.7\%} & 68.4\% & 53.4\% & 65.7\% & 68.8\% & 62.6\% \\
$\pi_0$ (fine-tuned)  & 72.7\% & 65.3\% & 38.3\% & 58.7\% & 75.2\% & 63.7\% & 25.6\% & 54.8\% \\
$\pi_0$-FAST (fine-tuned) & 75.3\% & 67.5\% & 42.9\% & 61.9\% & 77.6\% & 68.2\% & 31.3\% & 59.0\% \\
GR00T N1.5 (fine-tuned)  & 69.3\% & 68.7\% & 35.8\% & 52.4\% & 46.7\% & 62.9\% & 17.5\% & 43.7\% \\
SpatialVLA (zero-shot)  & 81.0\% & 69.6\% & 59.3\% & 70.0\% & 89.5\% & 71.7\% & 36.2\% & 65.8\% \\
SpatialVLA (fine-tuned)  & 86.0\% & 77.9\% & 57.4\% & 73.7\% & 88.0\% & 72.7\% & 41.8\% & 67.5\% \\
MolmoAct (zero-shot) & 71.3\% & 73.8\% & 66.5\% & 70.5\% & 57.8\% & 43.8\% & 76.7\% & 59.3\% \\
{MolmoAct (fine-tuned)} & {77.7\%} & {77.1\%} & {60.0\%} & {71.6\%} & {76.1\%} & {61.3\%} & \textbf{78.8\%} & \textbf{72.1\%} \\
\rowcolor{red!8}NORA-Long (zero-shot) & 74.2\% & 75.0\% & 31.7\% & 60.3\% &36.0\%	&73.0\%	&16.9\%& 42.0\%   \\
\rowcolor{blue!8} \model{}-FAST (zero-shot) & 79.5\% & \bf 90.9\% & 51.5\% & 74.0\% & 67.3\% & 71.6\% & 24.0\% & 54.3\% \\
\rowcolor{blue!8} \model{}-FAST (fine-tuned) & 88.6\% & 86.4\% & 41.2\% & 72.1\% & 85.2\% & \bf 85.2\% & 31.7\% & 67.4\% \\
\rowcolor{blue!8} \model{} (zero-shot) & 85.6\% & 88.6\% & 56.7\% & 76.9\% & 73.0\% & 80.1\% & 26.2\% & 59.7\% \\
\rowcolor{blue!8} \model{} (fine-tuned) & 92.8\% & 78.7\% & 62.2\% & 77.9\% & \bf 95.0\% & 75.7\% & 41.5\% & 70.7\% \\
\rowcolor{blue!8} \model{} (DPO) & \textbf{94.0\%}	& 88.0\%	& 66.4\% &	\textbf{82.8\%} & 92.6\% & 79.0\% & 44.1\% & 71.9\%\\
$\Delta$ from DPO & \textcolor{ForestGreen}{1.2\%}	& \textcolor{ForestGreen}{9.3\%}	& \textcolor{ForestGreen}{4.2\%} & \textcolor{ForestGreen}{4.9\%} & \textcolor{red}{-2.4\%} & \textcolor{ForestGreen}{3.3\%} & \textcolor{ForestGreen}{2.6\%} & \textcolor{ForestGreen}{1.2\%} \\
\bottomrule
\end{tabular}
 }
\end{table*}

As evident in \cref{tab:simplerenv-eval,tab:libero-eval,tab:real_robot_results}, \model{} generally outperforms all the baselines as analyzed below.

\paragraph{SimplerEnv.} The results in \cref{tab:simplerenv-eval} clearly show a superiority of \model{} in visual matching evaluation. Particularly on \emph{pick coke can} and \emph{move near} tasks, zero-shot \model{} outperform all the baseline zero-shot models by a wide margin of 4.6\% and 10.7\%, respectively. The performance advantage on these two tasks still holds for the fine-tuned variant of \model{} by 6.8\% and 0.8\%, respectively, against fine-tuned SpatialVLA. However, for \emph{open/close drawer} task this performance gain is not present. Magma far outperforms all the models in this regard, but its overall performance is far below even zero-shot \model{}. This could be attributed to its limited adaptation to such dragging and pushing actions that are relatively less prevalent in the pre-training dataset than pick and place actions---based on a keyword search on the task descriptions of Open X Embodiment pre-training dataset. However, the performance of fine-tuned \model{} on this task is still better by 4.8\% than the generally next-most capable fine-tuned model of SpatialVLA. For visual matching, overall zero-shot and fine-tuned variants of \model{} surpass the other equivalent next-best models by a wide 6.4\% and 4.2\%, respectively.

For variant aggregation setting, post-DPO \model{} performs comparably to the best model MolmoAct. However, the minute performance advantage of MolmoAct comes from a performance large performance advantage on \emph{drawer open/close} tasks. Whereas, it has huge underperformance on \emph{pick coke} and \emph{move next} tasks. Thus, \model{} could be considered more robust across tasks and variable visual settings.

\begin{table}[t]
\caption{Comparison of different baselines on spatial, object, goal, and long-horizon evaluation in LIBERO. The baseline results are taken from \citet{lee2025molmoactactionreasoningmodels}. Each subtask is evaluated across three random seed.}
\label{tab:libero-eval}
\centering
\resizebox{\linewidth}{!}{
\begin{tabular}{lcccc>{\columncolor{gray!10}}c}
\toprule
\textbf{Baseline} & \textbf{Spatial} & \textbf{Object} & \textbf{Goal} & \textbf{Long} & \textbf{Avg} \\
\midrule
TraceVLA  & 84.6\% & 85.2\% & 75.1\% & 54.1\% & 74.8\% \\
Octo-Base  & 78.9\% & 85.7\% & 84.6\% & 51.1\% & 75.1\% \\
OpenVLA  & 84.7\% & 88.4\% & 79.2\% & 53.7\% & 76.5\% \\
SpatialVLA  & 88.2\% & 89.9\% & 78.6\% & 55.5\% & 78.1\% \\
CoT-VLA  & 87.5\% & 91.6\% & 87.6\% & 69.0\% & 83.9\% \\
WorldVLA  & 87.6\% & 96.2\% & 83.4\% & 60.0\% & 79.1\% \\
$\pi_0$-FAST  & 96.4\% &  96.8\% & 88.6\% & 60.2\% & 85.5\% \\
$\pi_0$  & 96.8\% & \bf \textbf{98.8}\% & \textbf{95.8}\% & 85.2\% & 94.2\% \\
ThinkAct  & 88.3\% & 91.4\% & 87.1\% & 70.9\% & 84.4\% \\
{MolmoAct-7B-D} & {87.0\%} & {95.4\%} & {87.6\%} & {77.2\%} & {86.6\%} \\
\rowcolor{red!8}NORA  & 85.6\% & 89.4\% & 80.0\% & 63.0\% & 79.5\% \\
\rowcolor{red!8}NORA-Long & 92.2\% & 95.4\% &  89.4\% & 74.6\% & 87.9\%\\
\rowcolor{blue!8} \model{} & 97.3\% & 96.4\% & 94.5\% & 89.6\% & 94.5\%\\
\rowcolor{blue!8} \model{} (DPO) & \textbf{98.0\%} & 96.0\% &  95.4\% & \bf 90.5\% & \bf 95.0\% \\
$\Delta$ from DPO & \textcolor{ForestGreen}{0.7\%} & \textcolor{red}{-0.4\%} & \textcolor{ForestGreen}{0.9\%} & \textcolor{ForestGreen}{1.0\%} & \textcolor{ForestGreen}{0.6\%} \\
\bottomrule
\end{tabular}
}
\end{table}
\paragraph{NORA vs. \model{}.} 
\begin{figure}
    \centering
    \includegraphics[width=\linewidth]{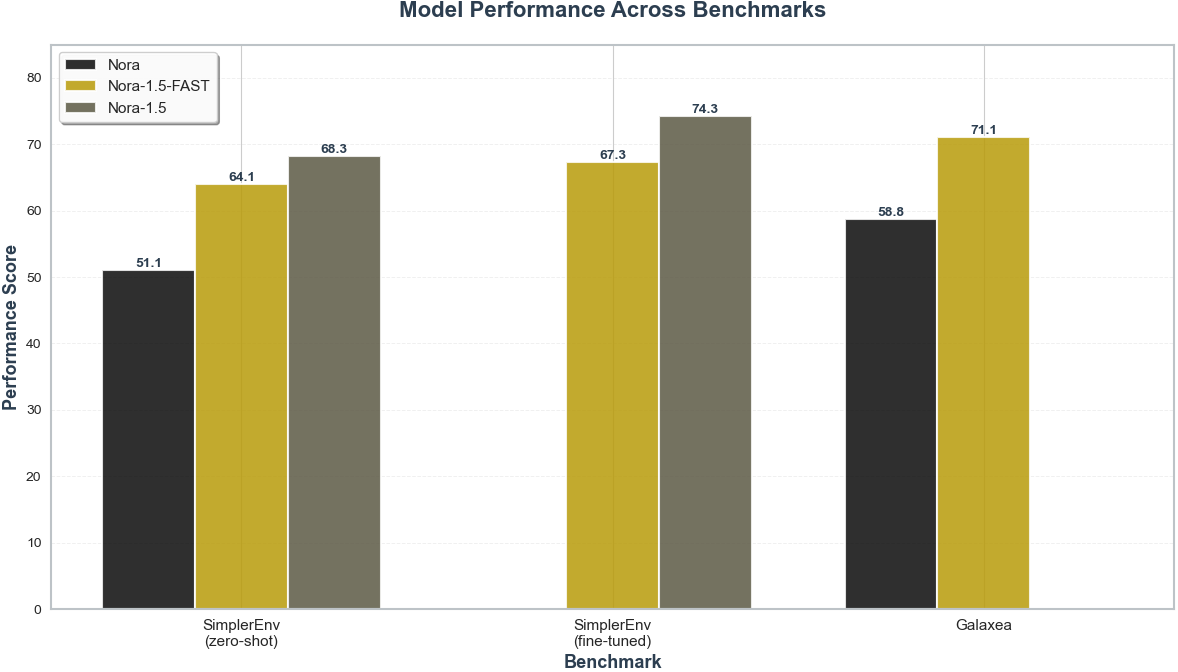}
    \caption{Comparing FAST+ with flow-matching.}
    \label{fig:nora-vs-nora1.5}
\end{figure}
We observe that \model{} consistently outperforms NORA across all benchmarks. According to \citet{intelligence2025pi05visionlanguageactionmodelopenworld}, flow matching in $\pi_{0.5}$ was primarily introduced to improve inference speed rather than performance. In contrast, in \model{}, coupling a flow-matching-based action expert with a pre-trained VLM-based autoregressive VLA leads to noticeable performance gains over the latter. We attribute this to the architectural design, where the flow-matching-based action expert and the autoregressive pre-trained VLA mutually benefit from each other. The flow-matching expert leverages the VLA’s rich representations—such as encoded observations, instructions, and overall plans—required for generating coherent actions. At the same time, the autoregressive VLA benefits from gradient feedback propagated through the action expert, enabling more effective learning such as improving the overall abstract plan for action expert to generate actions. In this way, the VLA is encouraged to plan the entire action trajectory that the expert subsequently leverages to generate actions. As shown in \cref{fig:nora-vs-nora1.5}, \model{}-FAST further surpasses NORA in both zero-shot and fine-tuned evaluations.
\paragraph{LIBERO.}
\cref{tab:libero-eval} shows that the performance gain by DPO  over the \model{} baselines is generally quite consistent and wide, except on LIBERO-Object evaluation tasks. This could be attributed to the limited variability of the object dimensions in these tasks, while the spatial setting and goal remain fixed. This lack of variability, as compared to the other tasks that differ in goals and spatial relationships, may have made these tasks easier. Thus, significant improvements could be harder to achieve. In fact, all the remaining models are not far behind the top models on these tasks, as compared to the remaining tasks. Overall, \model{} outperforms recent state-of-the-art models such as $\pi_0$.

\paragraph{Performance with Limited Real-life Robot Training Data.}

\begin{table*}
\centering
\caption{Experimental results of \model{} and baselines on nine real-world tasks with Galaxea A1 robotic arm. \textbf{Task Format} indicates the types of the physical objects---seen (\texttt{S}) vs unseen (\texttt{U})---and how they are related in the task. The \textcolor{red}{red}-inked objects are \textcolor{red}{distractors} (\textcolor{red}{Dist.}) in the setup. Succ. := \% success rate$_\tau$}
\resizebox{\textwidth}{!}{
\begin{tabular}{l p{5.5cm} c c c c c c c c c }
\toprule
\bf Task Format&\multirow{3}{*}{\bf Task ($\tau$)} & \multicolumn{3}{c}{$\pi_0$} & \multicolumn{3}{c}{NORA} & \multicolumn{3}{c}{\model{}-FAST}\\
Target(s) \textcolor{red}{[Distractor]} && \multicolumn{3}{c}{(3.3B)} & \multicolumn{3}{c}{(3B)} & \multicolumn{3}{c}{(3.3B)} \\
&& $\stackrel{\text{Part.}}{\text{Succ.}} \uparrow$ &  \textcolor{red}{Dist.} $\downarrow$ &  Succ. $\uparrow$ &  $\stackrel{\text{Part.}}{\text{Succ.}} \uparrow$ &  \textcolor{red}{Dist.} $\downarrow$ &  Succ. $\uparrow$ &  $\stackrel{\text{Part.}}{\text{Succ.}} \uparrow$ &  \textcolor{red}{Dist.} $\downarrow$ &  Succ. $\uparrow$ \\
\midrule
Put \texttt{U} in \texttt{U} & $\tau_1: $ Put eggplant in bowl & 90\% & - & 80\% & 90\% & - & 90\% & 100\% & - & \bf 100\% \\
& $\tau_2: $ Put apple in plate & 70\% & - & 30\% & 100\% & - & 80\% & 100\% & - & \bf 90\% \\
& $\tau_3: $ Put mango in basket & 90\% & - & 80\% & 80\% & - & 70\% & 90\% & - & \bf 90\% \\
\midrule
Put \texttt{U} in \texttt{S} \textcolor{red}{[\texttt{S}]}  & $\tau_4: $ Put strawberry in plate \textcolor{red}{[apple]}   & 0\% & 90\% & 0\% & 70\% & 0\% & \textbf{70}\% & 70\% & 10\% & \textbf{70}\% \\
&$\tau_5: $ Put grape in plate \textcolor{red}{[eggplant]}       & 0\% & 90\% & 0\% & 70\% & 20\% & 50\% & 80\% & 20\% & \bf 80\% \\
&$\tau_6: $ Put orange in plate \textcolor{red}{[banana]}       & 0\% & 100\% & 0\% & 30\% & 20\% & 40\% & 70\% & 30\% & \bf 60\% \\
\midrule
Move \texttt{U} to \texttt{U} \textcolor{red}{[\texttt{S}]} & $\tau_7:$ Move strawberry to banana \textcolor{red}{[apple]} & 50\% & 50\% & 10\% & 60\% & 20\% & 20\% & 60\% & 10\% & \bf 40\% \\
& $\tau_8: $ Move orange to banana \textcolor{red}{[apple]}       & 50\% & 30\% & 10\% & 80\% & 0\% & 50\% & 70\% & 20\% & \textbf{60}\% \\
& $\tau_9: $ Move cube to orange \textcolor{red}{[banana]}       & 50\% & 50\% & 20\% & 80\% & 20\% &  60\% & 70\% & 0\% & 50\% \\
\midrule
\multicolumn{2}{c}{Average} & 44.44\% & 68.33\% & 25.55\% & 73.3\% & \textbf{13.3\%} & 58.88\% & \textbf{78.88\%} & 15.00\% & \textbf{71.11\%} \\
\bottomrule
\end{tabular}
}
\label{tab:real_robot_results}
\end{table*}

\cref{tab:real_robot_results} shows the performance of $\pi_0$, NORA, and \model{} on the Galaxea A1 robotic arm. Across all experimental settings, \model{} outperforms these strong baselines by 13\% to 46\%. The improvement is larger for tasks with unseen distractors, suggesting the robustness of \model{}. Although we jointly optimize the flow-matching and autoregressive losses when fine-tuning the model on Galaxea A1 data, we observe that flow-matching-based action generation performs worse than autoregressive decoding. This differs from our observations of SimplerEnv and LIBERO (\cref{fig:nora-vs-nora1.5}), where flow-matching-based generation performs substantially better. We believe this difference arises from the smaller real-robot dataset (50K frames) as compared to SimplerEnv (4M frames). Since unlike $\pi_0$ our action expert lacks extensive flow-matching pre-training, it would likely require more data to effectively adapt than the autoregressive VLM backbone. This could explain the performance advantage of flow-matching-based generation on SimplerEnv and LIBERO, where larger fine-tuning datasets are available.

\subsection{Impact of DPO-based Post-training}
\begin{table}[ht!]
\caption{Ablation study on the proxy reward for DPO of \model{} through SimplerEnv evaluation. \texttt{WM} stands for world model-guided goal-based reward (\cref{eq:goal-re}) and \texttt{GTA} stands for the reward based on ground-truth action (\cref{eq:act-re}). VM:=Visual Matching, VA:=Variant Aggregation, PCC:=Pick Coke Can, MN:=Move Near, DR:=Open/Close Drawer.}
\label{tab:simplerenv-ablation}
\centering
\resizebox{\linewidth}{!}{
\begin{tabular}{lccc>{\columncolor{gray!10}}cccc>{\columncolor{gray!10}}c}
\toprule
\multirow{2}{*}{\textbf{Reward}} & 
\multicolumn{4}{c}{\textbf{VM}} & 
\multicolumn{4}{c}{\textbf{VA}} \\
\cmidrule(lr){2-5} \cmidrule(lr){6-9}
& PCC & MN & DR & Avg 
& PCC & MN & DR & Avg \\
\midrule
SFT (no reward) & 92.8\% & 78.7\% & 62.2\% & 77.9\% & 95.0\% & 75.7\% & 41.5\% & 70.7\% \\
\rowcolor[gray]{.9}\multicolumn{9}{c}{\textbf{Reward Techniques for DPO}}\\
\texttt{WM} (endgoal) & 93.6\% & 73.9\% & 61.6\% & 76.4\% & 95.1\% & 74.6\% & \bf 47.7\% & 72.5\% \\
\texttt{WM} (subgoal) & \bf 95.5\% & 81.8\% & 58.8\% & 78.7\% & \bf 95.3\% & \bf 80.7\% & 40.5\% & 72.2\% \\
\texttt{GTA} & 92.8\% & 86.0\% & 64.4\% & 81.2\% & 94.6\% & 80.1\% & 42.9 \% & 72.5\% \\
\texttt{WM} (endgoal) + \texttt{GTA} & 94.0\% & \bf 88.0\% & \bf 66.4\% &	\textbf{82.8\%} & 92.6\% & 79.0\% & 44.1\% & 71.9\%\\
\texttt{WM} (subgoal) + \texttt{GTA} & 92.4\% & 83.0\% & 61.6\% & 79.0\% & 93.4\% & 80.1\% & 45.4\% & \bf 73.0\% \\
\bottomrule
\end{tabular}
}
\end{table}
We study the impact of three reward formulations: an action-based reward (\cref{eq:act-re}), WM-guided goal-based reward (\cref{eq:goal-re}), and a linear combination of these two (\cref{eq:tot-re}). We perform these experiments primarily on the SimplerEnv and LIBERO simulation tasks.

\paragraph{SimplerEnv.}
The results in \cref{tab:simplerenv-ablation} suggest that different reward strategies utilized for DPO generally outperform the SFT baseline for both Visual Matching and Variant Aggregation. The WM-guided goal-based reward proved to be an exception, leading to a performance degradation on the ``Move Near'' task for both visual matching and variant aggregation. This is likely because a purely goal-directed guidance does not account for the implicit safety constraints of the task. Based on the success criteria of ``Move Near'' in SimplerEnv, the robot is required to avoid obstacles while completing the task. Notably, in \cref{tab:simplerenv-ablation}, the hybrid \texttt{WM} + \texttt{GTA} model achieves the best overall performance of 82.8\% on Visual matching and a significant increase on the ``Move Near'' subtask, suggesting that by adding ground truth action to the goal-based reward, 

\paragraph{LIBERO.}
The performance on LIBERO also improves with DPO, although the gains are smaller as compared to SimplerEnv. This is likely because the SFT model already performs strongly on LIBERO, leaving limited room for further improvement. Among the LIBERO benchmarks, the most challenging task is LIBERO-Long, where we observe consistent performance gains of 1\% to 1.7\% across all reward modeling techniques.
\begin{table*}[t]
\centering
\caption{Comparing \model{}-FAST w/ and w/o DPO. Metrics shown separately as Correct/Dist./Succ. The newly introduced tasks are : $\tau_{10}:$ Put mango in plate \textcolor{red}{[apple, grape]}, $\tau_{11}:$ Put mango in plate \textcolor{red}{[apple, grape, orange]}, $\tau_{12}:$ Put orange in plate \textcolor{red}{[apple, grape]}, $\tau_{13}:$ Put orange in plate \textcolor{red}{[apple, grape, mango]}. Remaining tasks can be found in \cref{tab:real_robot_results}.}
\resizebox{\linewidth}{!}{
\begin{tabular}{@{}l l ccc ccc ccc ccc@{}}
\toprule
Format & \multirow{3}{*}{Task ($\tau$)} & \multicolumn{9}{c}{Reward Method for DPO} \\
\cmidrule(lr){3-14}
Target(s) \textcolor{red}{[Dist.]} & & \multicolumn{3}{c}{\model{}-FAST} & \multicolumn{3}{c}{w/ \texttt{WM} (subgoal)+\texttt{GTA}} & \multicolumn{3}{c}{w/ \texttt{WM} (endgoal)+\texttt{GTA}} & \multicolumn{3}{c}{w/ \texttt{GTA}} \\
\cmidrule(lr){3-5}\cmidrule(lr){6-8}\cmidrule(lr){9-11}\cmidrule(lr){12-14}
 & & $\stackrel{\text{Part.}}{\text{Succ.}}\uparrow$ & \textcolor{red}{Dist.}$\downarrow$ & Succ.$\uparrow$ & $\stackrel{\text{Part.}}{\text{Succ.}}\uparrow$ & \textcolor{red}{Dist.}$\downarrow$ & Succ.$\uparrow$ & $\stackrel{\text{Part.}}{\text{Succ.}}\uparrow$ & \textcolor{red}{Dist.}$\downarrow$ & Succ.$\uparrow$ & $\stackrel{\text{Part.}}{\text{Succ.}}\uparrow$ & \textcolor{red}{Dist.}$\downarrow$ & Succ.$\uparrow$  \\
\midrule
Put \texttt{U} in \texttt{U} & $\tau_1 $ & \cellcolor{gray!12}100 & \cellcolor{gray!12}- & \cellcolor{gray!12}100 & \cellcolor{gray!12}100 & \cellcolor{gray!12}- & \cellcolor{gray!12}100 & \cellcolor{gray!12}90 & \cellcolor{gray!12}- & \cellcolor{gray!12}90 & \cellcolor{gray!12}100 & \cellcolor{gray!12}- & \cellcolor{gray!12}100  \\
& $\tau_2 $ & \cellcolor{gray!12}100 & \cellcolor{gray!12}- & \cellcolor{gray!12}90 & \cellcolor{gray!12}100 & \cellcolor{gray!12}- & \cellcolor{gray!12}100 & \cellcolor{gray!12}100 & \cellcolor{gray!12}- & \cellcolor{gray!12}100 & \cellcolor{gray!12}90 & \cellcolor{gray!12}- & \cellcolor{gray!12}90 \\
& $\tau_3 $ & \cellcolor{gray!12}90 & \cellcolor{gray!12}- & \cellcolor{gray!12}90 & \cellcolor{gray!12}90 & \cellcolor{gray!12}- & \cellcolor{gray!12}90 & \cellcolor{gray!12}90 & \cellcolor{gray!12}- & \cellcolor{gray!12}90 & \cellcolor{gray!12}90 & \cellcolor{gray!12}- & \cellcolor{gray!12}90  \\
\midrule
Put \texttt{U} in \texttt{S} \textcolor{red}{[\texttt{S}]}  & $\tau_4 $  & \cellcolor{gray!12}70 & \cellcolor{gray!12}10 & \cellcolor{gray!12}70 & \cellcolor{gray!12}70 & \cellcolor{gray!12}0 & \cellcolor{gray!12}70 & \cellcolor{gray!12}70 & \cellcolor{gray!12}0 & \cellcolor{gray!12}70 & \cellcolor{gray!12}70 & \cellcolor{gray!12}0 & \cellcolor{gray!12}70  \\
& $\tau_5 $        & \cellcolor{gray!12}80 & \cellcolor{gray!12}20 & \cellcolor{gray!12}80 & \cellcolor{gray!12}80 & \cellcolor{gray!12}10 & \cellcolor{gray!12}80 & \cellcolor{gray!12}90 & \cellcolor{gray!12}0 & \cellcolor{gray!12}70 & \cellcolor{gray!12}80 & \cellcolor{gray!12}10 & \cellcolor{gray!12}60  \\
& $\tau_6 $ & \cellcolor{gray!12}70 & \cellcolor{gray!12}30 & \cellcolor{gray!12}60 & \cellcolor{gray!12}90 & \cellcolor{gray!12}0 & \cellcolor{gray!12}80 & \cellcolor{gray!12}90 & \cellcolor{gray!12}0 & \cellcolor{gray!12}80 & \cellcolor{gray!12}70 & \cellcolor{gray!12}0 & \cellcolor{gray!12}70  \\
& $\tau_{10} $ & \cellcolor{gray!12}80 & \cellcolor{gray!12}10 & \cellcolor{gray!12}70 & \cellcolor{gray!12}90 & \cellcolor{gray!12}10 & \cellcolor{gray!12}90 & \cellcolor{gray!12}70 & \cellcolor{gray!12}10 & \cellcolor{gray!12}70 & \cellcolor{gray!12} 90 &\cellcolor{gray!12} 0 &  \cellcolor{gray!12} 70 \\
& $\tau_{11}$ & \cellcolor{gray!12}40 & \cellcolor{gray!12}20 & \cellcolor{gray!12}10 & \cellcolor{gray!12}50 & \cellcolor{gray!12}50 & \cellcolor{gray!12}30 & \cellcolor{gray!12}40 & \cellcolor{gray!12}40 & \cellcolor{gray!12}40 & \cellcolor{gray!12} 50 & \cellcolor{gray!12} 40 & \cellcolor{gray!12} 10\\
& $\tau_{12}$ & \cellcolor{gray!12}60 & \cellcolor{gray!12}10 & \cellcolor{gray!12}10 & \cellcolor{gray!12}80 & \cellcolor{gray!12}10 & \cellcolor{gray!12}40 & \cellcolor{gray!12}50 & \cellcolor{gray!12}0 & \cellcolor{gray!12}10 & \cellcolor{gray!12}40 & \cellcolor{gray!12}30 & \cellcolor{gray!12}0 \\
& $\tau_{13}$ & \cellcolor{gray!12}50 & \cellcolor{gray!12}20 & \cellcolor{gray!12}10 & \cellcolor{gray!12}80 & \cellcolor{gray!12}20 & \cellcolor{gray!12}30 & \cellcolor{gray!12}30 & \cellcolor{gray!12}20 & \cellcolor{gray!12}10 & \cellcolor{gray!12}50 & \cellcolor{gray!12}0 & \cellcolor{gray!12}0\\
\midrule
Move \texttt{U} to \texttt{U} \textcolor{red}{[\texttt{S}]} & $\tau_{7}$ & \cellcolor{gray!12}60 & \cellcolor{gray!12}0 & \cellcolor{gray!12}40 & \cellcolor{gray!12}80 & \cellcolor{gray!12}10 & \cellcolor{gray!12}70 & \cellcolor{gray!12}80 & \cellcolor{gray!12}10 & \cellcolor{gray!12}60 & \cellcolor{gray!12}80 & \cellcolor{gray!12}10 & \cellcolor{gray!12}70  \\
& $\tau_{8}$ & \cellcolor{gray!12}70 & \cellcolor{gray!12}20 & \cellcolor{gray!12}50 & \cellcolor{gray!12}100 & \cellcolor{gray!12}0 & \cellcolor{gray!12}70 & \cellcolor{gray!12}90 & \cellcolor{gray!12}0 & \cellcolor{gray!12}60 & \cellcolor{gray!12}70 & \cellcolor{gray!12}20 & \cellcolor{gray!12}60  \\
& $\tau_{9}$ & \cellcolor{gray!12}70 & \cellcolor{gray!12}20 & \cellcolor{gray!12}60 & \cellcolor{gray!12}80 & \cellcolor{gray!12}10 & \cellcolor{gray!12}60 & \cellcolor{gray!12}60 & \cellcolor{gray!12}20 & \cellcolor{gray!12}50 & \cellcolor{gray!12}80 & \cellcolor{gray!12}20 & \cellcolor{gray!12}60 \\
\midrule
\multicolumn{2}{c}{Average} & \cellcolor{gray!12}72.30 & \cellcolor{gray!12}16.00 & \cellcolor{gray!12}56.92 & \cellcolor{gray!12}\textbf{83.84} & \cellcolor{gray!12}12.00 & \cellcolor{gray!12}\textbf{70.00} & \cellcolor{gray!12}73.07 & \cellcolor{gray!12}\textbf{10.00} & \cellcolor{gray!12}61.53  &\cellcolor{gray!12} 73.84 &\cellcolor{gray!12} 13 & \cellcolor{gray!12}57.69\\
\multicolumn{2}{c}{(Improvement)} & & & & \cellcolor{green!12}11.54 & \cellcolor{green!12}4.00 & \cellcolor{green!12}13.08 & \cellcolor{green!12}0.77 & \cellcolor{green!12}6.00 & \cellcolor{green!12}4.61 &\cellcolor{green!12} 1.54 &\cellcolor{green!12} 3.00 &\cellcolor{green!12} 0.77\\
\multicolumn{2}{c}{Average (Unseen)} & \cellcolor{gray!12}65.00 & \cellcolor{gray!12}16.00 & \cellcolor{gray!12}46.00 & \cellcolor{gray!12}\textbf{80.00} & \cellcolor{gray!12}12.00 & \cellcolor{gray!12}\textbf{62.00} & \cellcolor{gray!12}67.00 & \cellcolor{gray!12}\textbf{10.00} & \cellcolor{gray!12}\textbf{52.00} & \cellcolor{gray!12} 68.00 & \cellcolor{gray!12} 13.00 & \cellcolor{gray!12} 47.00 \\
\multicolumn{2}{c}{(Improvement)} & & & & \cellcolor{green!12}15.00 & \cellcolor{green!12}4.00 & \cellcolor{green!12}16.08 & \cellcolor{green!12}2.00 & \cellcolor{green!12}6.00 & \cellcolor{green!12}6.00 &\cellcolor{green!12}3.00 &\cellcolor{green!12}3.00&\cellcolor{green!12}1.00\\
\bottomrule
\end{tabular}
}
\label{tab:real_robot_results2}
\end{table*}

\paragraph{Galaxea A1 Robotic Arm.}
After observing marginal to substantial improvements from our action-rewarding strategies and DPO across simulation benchmarks such as LIBERO and SimplerEnv, we next evaluate whether these gains carry over to real-robot settings. Real-world experiments also allow us to understand more concretely how DPO-based post-training benefits VLA models. To this end, we extend the ``Put \texttt{U} in \texttt{S} \textcolor{red}{[\texttt{S}]}'' task suite by adding four more variants with additional distractors to increase task difficulty.

As shown in \cref{tab:real_robot_results2}, across all thirteen tasks, the DPO-trained \model{} achieves a notable 13\% performance improvement. Specifically, correct-object grasping accuracy increases by 11\%, while unintended grasps of distractors decrease by 4\%. While DPO marginally improves the performance for in-domain or seen objects/tasks, we see a larger performance improvement of 15\%-16\% on unseen tasks and objects. These trends indicate that DPO contributes primarily in two ways: (1) enhancing the affordance and reliability of the grasping action, and (2) improving the model’s ability to focus on the intended target object.

\begin{figure}[t]
    \centering
    \begin{subfigure}[t]{0.48\linewidth}
        \centering
        \includegraphics[width=\linewidth]{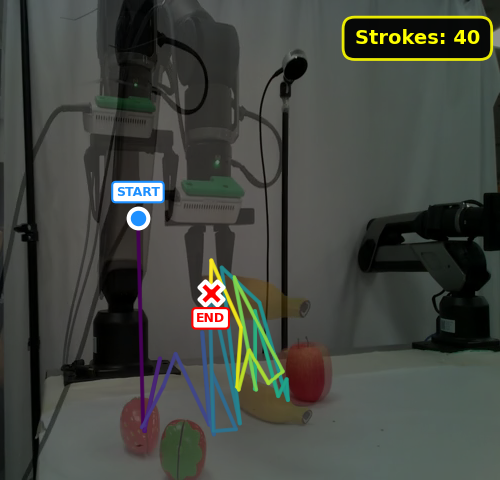}
        \caption{\model{} without DPO. The gripper trajectory exhibits frequent fixations and zig-zag motions, often resulting in failed grasps and grasp attempts toward distractor objects.}
        \label{fig:affordance:nodpo}
    \end{subfigure}
    \hfill
    \begin{subfigure}[t]{0.48\linewidth}
        \centering
        \includegraphics[width=\linewidth]{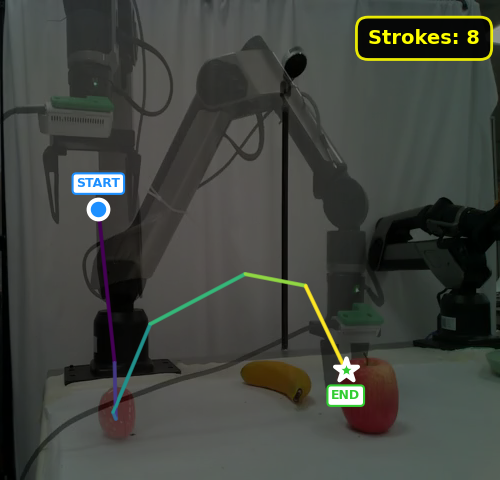}
        \caption{\model{} with reward-driven DPO post-training. The gripper trajectory becomes smoother and more consistent, with
        fewer corrective strokes and more reliable target grasps.}
        \label{fig:affordance:dpo}
    \end{subfigure}
    \caption{
        Effect of DPO post-training on real-robot gripper trajectories
        for the Galaxea A1 arm. Compared to the non-DPO baseline
        (a), the DPO-trained \model{}
        (b) executes smoother trajectories
        with fewer strokes, aligning with the reduced number of action
        chunks and improved grasp success reported in
        \cref{tab:real_robot_results2}.
    }
    \label{fig:affordance}
\end{figure}

\cref{fig:affordance} further illustrates these effects. With reward-driven DPO post-training, the gripper trajectory becomes more consistent and well-formed, whereas the non-DPO model exhibits more fixations and zig-zag motions. Consequently, while \model{} with DPO requires only 7.0 action chunks on average to grasp the target, \model{} without DPO takes 9.7 action chunks, indicating that the latter struggles to execute smooth and efficient grasps—an issue mitigated by DPO post-training. The results also show that DPO helps reduce the likelihood of the robot accidentally picking up distractor objects.

\subsection{Ablations}
\label{sec:ablation}

\begin{table}[t]
\caption{Ablation study on the proxy reward for DPO of \model{} through spatial, object, goal, and long-horizon evaluation in LIBERO. \texttt{WM} stands for world model-guided goal-based reward (\cref{eq:goal-re}) and \texttt{GTA} stands for the reward based on ground-truth action (\cref{eq:act-re}).}
\label{tab:libero-ablation}
\centering
\resizebox{\linewidth}{!}{
\begin{tabular}{lcccc>{\columncolor{gray!10}}c}
\toprule
\textbf{Reward} & \textbf{Spatial} & \textbf{Object} & \textbf{Goal} & \textbf{Long} & \textbf{Avg} \\
\midrule
SFT (no reward) & 97.3\% & \bf 96.4\% & 94.5\% & 89.6\% & 94.5\%\\
\rowcolor[gray]{.9}\multicolumn{6}{c}{\textbf{Reward Techniques for DPO}}\\
\texttt{WM} (endgoal) & 98.0\% & 96.0\% & \bf 95.4\% & 90.5\% & \bf 95.0\% \\
\texttt{GTA} & \bf 98.3\% & 95.9\% & 94.7\% & 90.7\% & 94.9\% \\
\texttt{WM} (endgoal) + \texttt{GTA} & 97.9\% & 95.9\% & 94.1\% & \bf 91.3\% & 94.8\% \\
\bottomrule
\end{tabular}
}
\end{table}

To determine the individual contribution of the various reward elements---WM-guided goal-based reward (\cref{eq:goal-re}) and ground-truth action-based reward (\cref{eq:act-re})---, we also curate preference datasets for SimplerEnv and LIBERO with the individual elementary rewards and apply DPO to the pre-trained \model{}. Evaluating the elementary rewards in SimplerEnv, as shown in \cref{tab:simplerenv-ablation}, reveals a general superiority of the combined reward (\cref{eq:tot-re}) for visual matching. Interestingly, the end-goal-based reward \texttt{WM} (endgoal) causes the performance on \emph{move near} and \emph{drawer open/close} tasks to drop below SFT. On the other hand, the subgoal-based reward \texttt{WM} (subgoal) is overall 1.7\% better than \texttt{WM} (endgoal) and overall beats the SFT baseline by a small margin. Specifically, \texttt{WM} (subgoal) surpasses the SFT baseline on \emph{move near} task by 3.1\%, whereas \texttt{WM} (endgoal) lags behind by 4.8\%. This could be indicative of the noisiness of the signal from the world model, where the guidance of the final goal image is noisier than the immediate subgoal images due to shaky long-term dependency modeling.

The ground-truth action-based reward (\texttt{GTA}) is generally superior to all other elementary rewards for visual matching. This reward might teach the model to follow the most straightforward trajectory to achieve the goal, achieving superior results. However, for \emph{pick coke can} task, this reward fails to surpass the SFT baseline and falls behind the other two elementary goal-based rewards. This may indicate the drawback of such a straightforward approach, which may induce certain biases in the model that may not work out in very specific cases.

For the evaluation with variant aggregation, the overall performance of all the elementary rewards are in the same ballpark. \texttt{WM} (subgoal) beats the other two elementary rewards on \emph{pick coke can} and \emph{move next} tasks by a small margin. In fact, all these elementary rewards beat the combined reward \texttt{WM} (endgoal) + \texttt{GTA} across all the tasks, but the overall performance is very comparable. Interestingly, the \texttt{WM} (subgoal) + \texttt{GTA} combination shows the most stable performance under this setting by outperforming all on average despite not being the best at any individual task. This may underscore the robustness of short-term subgoal modeling to the changing visuals.

For real-world Galaxea A1-based experiments (\cref{tab:real_robot_results2}), both the \texttt{WM} (subgoal)+\texttt{GTA} and \texttt{WM} (endgoal)+\texttt{GTA} reward methods improve performance over the SFT model. However, \texttt{WM} (subgoal)+\texttt{GTA} reward consistently yields stronger gains across all metrics—including grasping the correct object, avoiding distractors, and overall task completion. A plausible explanation is the presence of unseen objects and varying environmental conditions in the real-robot setup. In such settings, localized guidance from subgoal rewards may offer a more reliable and less noisy training signal than endgoal-based rewards. We also observe that \texttt{GTA} rewards offer limited benefits for real robots. In fact, for most tasks in the ``Put~\texttt{U}~in~\texttt{S}'' category—including all newly introduced challenging tasks—its performance is worse than the baseline. This reinforces our assumption that, in real-world environments, multiple trajectories can successfully accomplish the same task. As a result, forcing the model to consistently follow a single labeled trajectory may introduce unnecessary noise into the robot's behavior when operating in unseen scenarios. By combining \texttt{subgoal/goal} information with \texttt{GTA} to construct the training dataset, we provide additional contextual signals that help guide the robot toward selecting an appropriate trajectory to complete the task.

For LIBERO, \cref{tab:libero-ablation} shows that the overall performance gain over SFT by both elementary and combined reward is quite minute. This could be ascribed to the diminished gain potential for LIBERO due to already high performance of the SFT baseline.

\section{Conclusion}

Increasing research and commercial interest in VLA models call for effective adaptation of these models to a wide-range of embodiments/robots. This work shows that preference optimization-based post-training improves adaptation of \model{} for both real world and simulation, given appropriate reward modeling. The experiments and analyses substantiate that our world model-driven goal- and action-based reward is quite potent proxy reward for DPO of our \model{}, resulting in significant performance gains. Our real-world evaluation also highlight the performance advantage of \model{} over the state-of-the-art open VLA model $\pi_0$. We hope this paper provides a sturdy foundation for the future research on post-training VLA models and for embodied AI as a whole.


{
    \small
    \bibliographystyle{ieeenat_fullname}
    \bibliography{aaai2026,custom}
}

\clearpage

\input{appendix}

\end{document}

%% file: appendix.tex
\appendix
\setcounter{page}{1}

\input{benchmarks}
\input{related-works}

\begin{figure}[ht]
\centering
\begin{subfigure}{\linewidth}
    \centering
    \includegraphics[width=\linewidth]{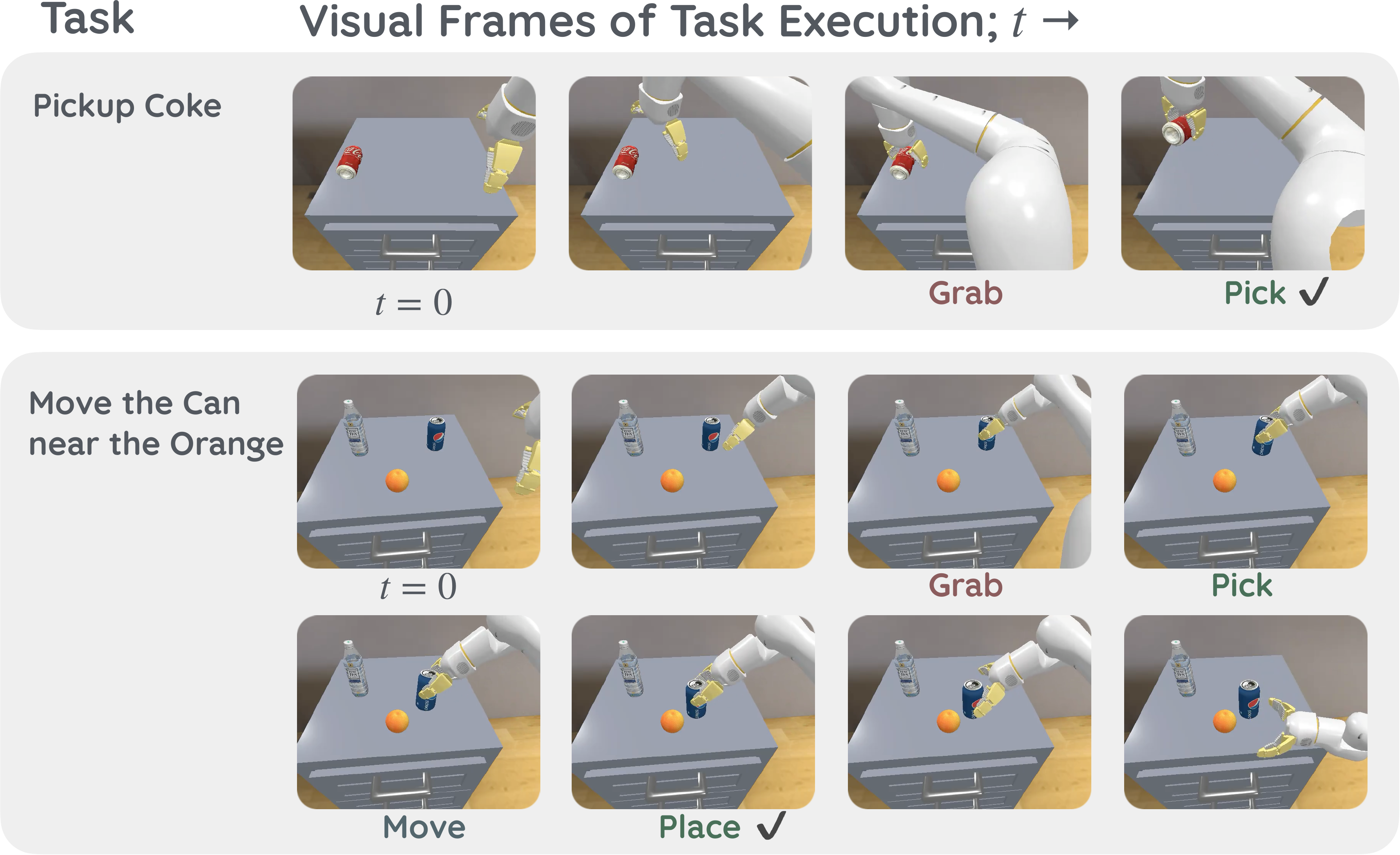}
    \caption{}
\end{subfigure}

\begin{subfigure}{\linewidth}
    \centering
    \includegraphics[width=\linewidth]{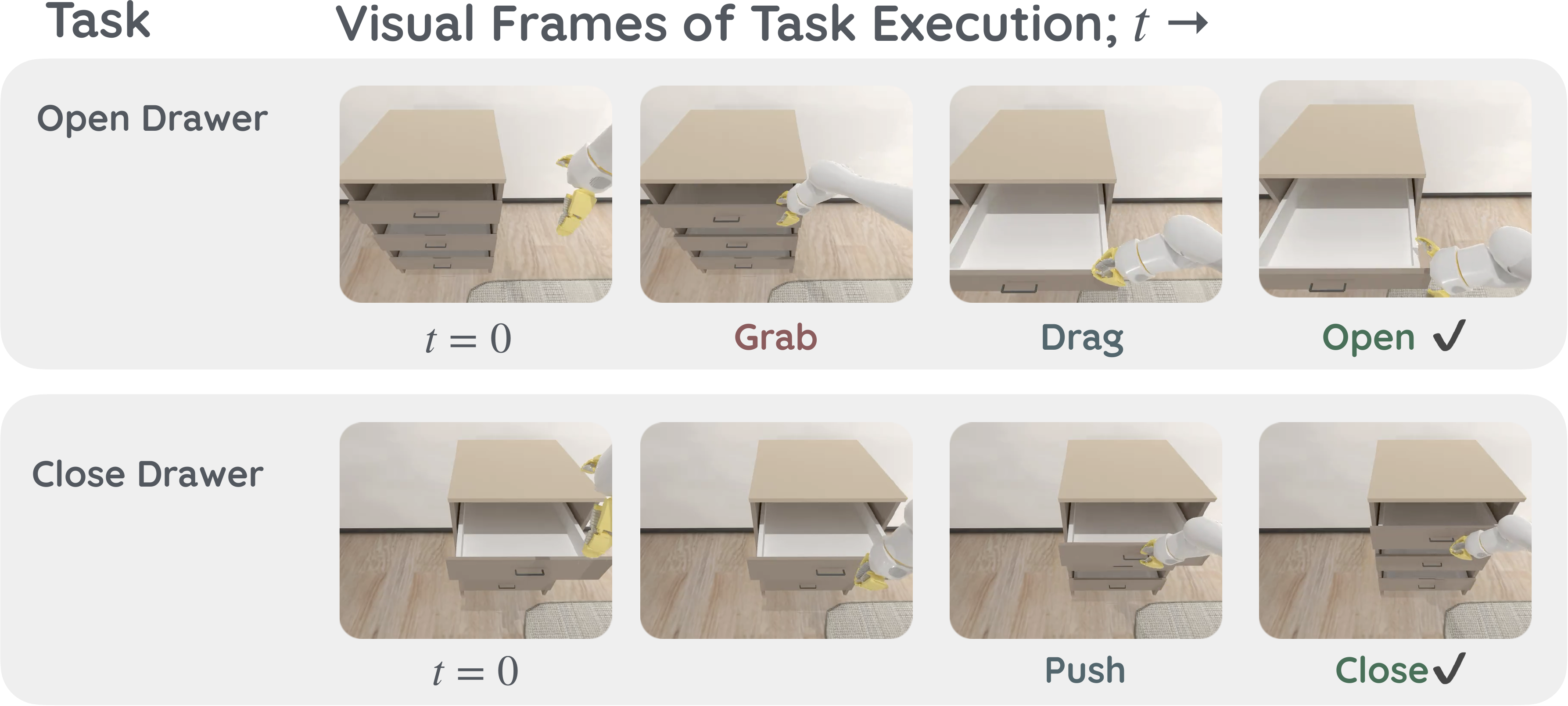}
    \caption{}
\end{subfigure}
\caption{Examples of \model{} executing evaluation tasks in SimplerEnv: (a) pickup coke and move object near another object and (b) open and close drawer.}
\label{fig:sim-exec}
\end{figure}
\begin{figure}
    \centering
    \includegraphics[width=\linewidth]{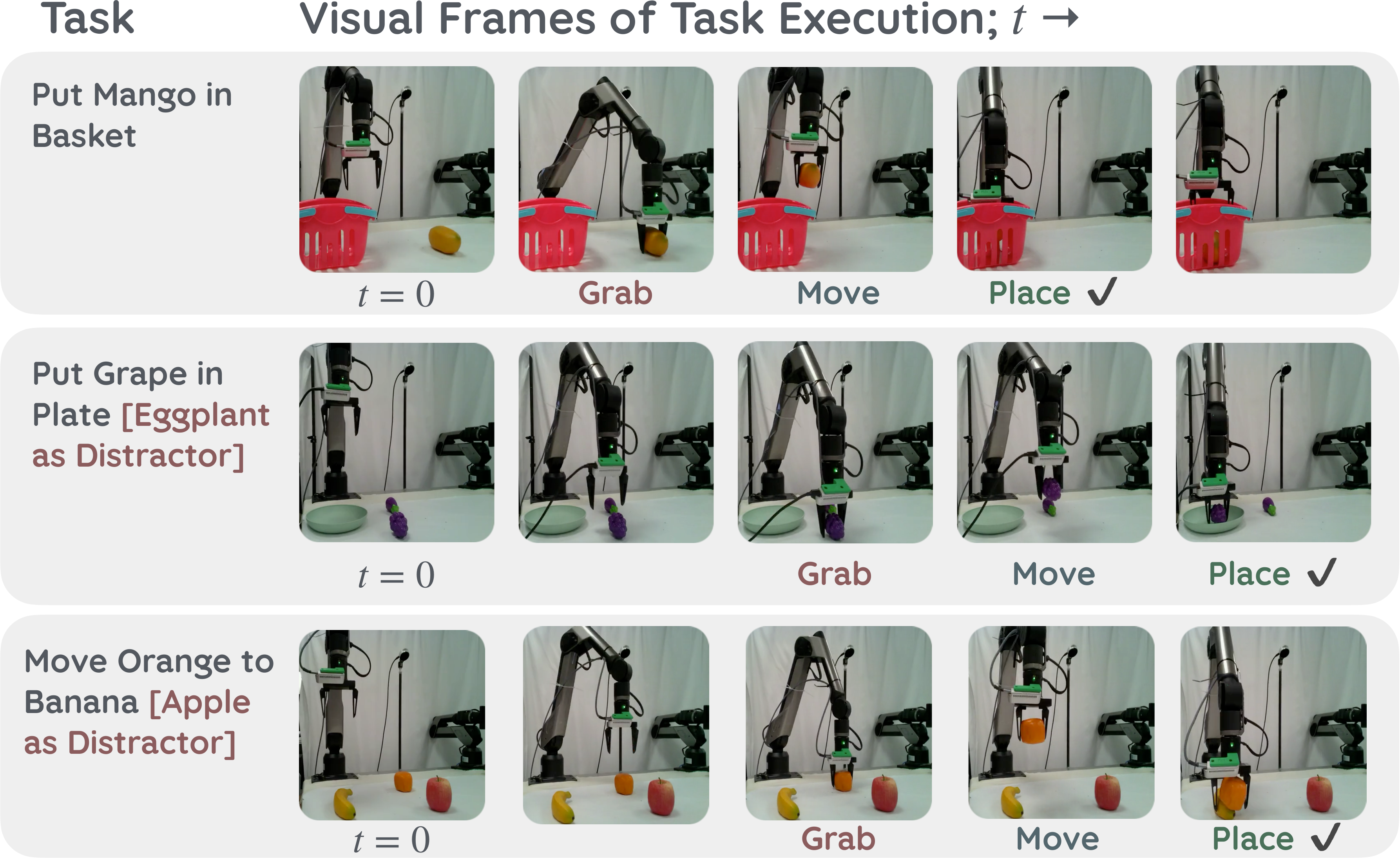}
    \caption{Examples of \model{} executing evaluation tasks with Galaxea A1 robotic arm in the real world.}
    \label{fig:galaxea-exec}
\end{figure}
\section{Model Architecture and Training Details}

\model{} is adapted from NORA-Long by initializing a new action expert that has approximately 400 million parameters. Embeddings of Qwen 2.5 VL and the newly initialized action expert interact through self-attention only. To prevent action representation of FAST token leaking to the action expert, the action expert is only allowed to attend embeddings corresponding to the language instruction and the image. During training, we optimize a joint cross entropy loss on FAST tokens as well as flow matching loss on the action expert: $\mathcal{L}_{\text{Loss}} = \mathcal{L}_{\text{CE}}+\alpha\mathcal{L}_{\text{FM}}
$ where $\mathcal{L}_{\text{CE}}$ denotes the cross entropy loss on FAST tokens, $\mathcal{L}_{\text{FM}}$ is same as Equation \ref{eq:fm} and $\alpha$ is a scaling term. We set $\alpha=10$ during our training.

\model{} is trained on the exact same subset of the Open-X-Embodiment dataset as NORA-Long. Previously, NORA-Long was trained for 900k gradient steps with a global batch size of 256. After initializing the action expert, we jointly optimize $\mathcal{L}_{\text{Loss}}$ for another 150k gradient steps with a global batch size of 512. We use a maximum learning rate of  $5e-5$, a linear warm up of $5000$ steps, and a cosine decay to $0$. We used a single node of H100 GPU to train this for 5 days, approximately using 960 H100 hours.

\section{Galaxea Data Collection}
We collected 1000 episodes of simple pick and place tasks via teleportation for fine-tuning $\pi_0$, \model{} and NORA-Long on Galaxea A1 robot. During the collection of data, we randomly place objects on the table and do not follow any order. We collected a total of 9 different tasks where each task has about \~100+ episodes.

\section{Baselines}

We use the following baselines for a comparative evaluation of our approaches.

\textbf{OpenVLA} \citep{kim2024openvla}: A VLA model is built upon a Llama 2 language model\cite{touvron2023llama} combined with a visual encoder that integrates pretrained features from DINOv2 \cite{dinov2} and SigLIP\cite{zhai2023sigmoid}. It is pretrained on the Open-X-Embodiment dataset~\citep{open_x_embodiment_rt_x_2023}, which comprises 970k real-world robot demonstrations.

\textbf{SpatialVLA} \citep{spatialvla2025}: A VLA model focused on spatial understanding for robot manipulation, incorporating 3D information such as spatial movement. It learns a generalist policy for spatial manipulation across diverse robots and tasks. SpatialVLA predicts four actions at a time.

\textbf{TraceVLA} \citep{tracevla2024}: A VLA model enhancing spatial-temporal reasoning via visual trace prompting. Built by fine-tuning OpenVLA on robot manipulation trajectories, it encodes state-action history as visual prompts to improve manipulation performance in interactive tasks.

\textbf{RT-1} \citep{brohan2023rt1roboticstransformerrealworld}: A scalable Robotics Transformer model designed to transfer knowledge from large task-agnostic datasets. Trained on diverse robotic data, RT-1 achieves a high level of generalization and task-specific performance across a variety of robotic tasks, demonstrating the value of open-ended task-agnostic training of high-capacity models.

\textbf{HPT}~\cite{wang2024hpt}.  
Heterogeneous Pre-trained Transformers (HPT) pretrain a shared
transformer trunk on a large mixture of heterogeneous robot and
video datasets, aligning proprioceptive and visual inputs into a
unified token sequence. The resulting policy improves generalization
across embodiments and tasks, and we use the released HPT policies
as SimplerEnv baselines.

\textbf{Octo-Base}~\cite{ghosh2024octo}.  
Octo is a transformer-based diffusion policy trained on
$\sim$800k trajectories from Open X-Embodiment. We use the
Octo-Base variant, a ViT-B–sized model that supports flexible
action and observation spaces and can be fine-tuned efficiently
for new robot setups.

\textbf{RoboVLM}~\cite{li2024robovlm}.  
RoboVLM is a framework for systematically studying design choices
in VLAs and building generalist policies from diverse VLM backbones,
architectures, and cross-embodiment data. We adopt their best-performing
RoboVLM policy as a strong generalist VLA baseline.

\textbf{$\pi_0$ and $\pi_0$-FAST}~\cite{black2024pi0}.  
$\pi_0$ is a vision-language-action model that attaches a
flow-matching action expert to a pretrained VLM and is trained on
a large cross-embodiment dataset for high-frequency, dexterous control.
$\pi_0$-FAST tokenize actions as discrete token using the FAST  tokenizer.  This enables faster convergence with lesser training compute. Both models serve as powerful generalist baselines.

\textbf{MolmoAct / MolmoAct-7B-D}~\cite{lee2025molmoact}.  
MolmoAct is an action reasoning VLA that factors control into
three stages: depth-aware perception tokens, mid-level spatial
trajectory traces, and low-level actions. We use the 7B-D variant,
MolmoAct-7B-D, which achieves strong zero-shot and fine-tuned
performance on SimplerEnv and LIBERO.

\textbf{Emma-X}~\cite{sun2025emmax}.  
Emma-X is a 7B VLA obtained by
fine-tuning OpenVLA on a hierarchical dataset derived from BridgeV2, with grounded chain-of-thought reasoning and look-ahead spatial guidance. 

\textbf{Magma}~\cite{yang2025magma}.  
Magma is a multimodal agentic foundation model that unifies vision,
language, and action for both digital UI navigation and physical
robot manipulation. It introduces visual planning traces and serves
as a large-scale generalist baseline in our real-robot comparisons.

\textbf{GR00T N1.5}~\cite{bjorck2025gr00tn1,nvidia2025gr00tn15}.  
GR00T N1 is an open VLA foundation model for humanoid robots with a
dual-system design: a vision-language backbone and a diffusion-based
action policy. GR00T N1.5 is an improved release with architectural
and data updates; we use the 3B N1.5 policy as a strong generalist
baseline.

\textbf{CoT-VLA}~\cite{zhao2025cotvla}.  
CoT-VLA augments VLAs with \emph{visual chain-of-thought} reasoning:
it first predicts subgoal images as visual plans and then generates
short action sequences to reach those subgoals, improving performance
on long-horizon and multi-step manipulation.

\textbf{WorldVLA}~\cite{cen2025worldvla}.  
WorldVLA unifies a VLA policy and an image world model in a single
autoregressive transformer, jointly modeling images, language, and
actions. The world model predicts future images conditioned on actions,
and the action head benefits from world-model feedback for better
planning.

\textbf{ThinkAct}~\cite{huang2025thinkact}.  
ThinkAct is a dual-system VLA that separates high-level reasoning
from low-level action. A multimodal LLM produces structured embodied
plans which are compressed into a visual latent, conditioning a
downstream action policy for few-shot adaptation and long-horizon
control.

\textbf{NORA and NORA-Long}~\cite{hung2025nora}.  
NORA is a 3B VLA built on Qwen2.5-VL-3B and trained on
Open X-Embodiment data with FAST tokenizer, designed to
provide strong performance under tight compute budgets. NORA-Long is a variant with an extended action horizon and the original NORA VLA.

%% file: benchmarks.tex
\section{Evaluation Settings and Metrics}

The comparative evaluation of our VLA model is performed under both real-world and simulated settings. On one hand, Galaxea A1 robotic arm is chosen as the embodiment for the real-world evaluation. On the other hand, LIBERO~\cite{liu2023libero} and SimplerEnv~\cite{li24simpler} simulated benchmarks are used to evaluate the VLA models under a diverse range of settings.

\paragraph{LIBERO~\cite{liu2023libero}} simulated benchmark comprising four subsets to test generalization across spatial layouts (LIBERO-Spatial), objects (LIBERO-Object), task goals (LIBERO-Goal), and long horizon tasks (LIBERO-Long). We followed the approaches of \citet{kim2024openvla} and  purged all the no-op actions during fine-tuning. For fine-tuning, we combined the data from four distinct subsets to train a single model. For evaluation, each of the four corresponding tasks was evaluated across 500 episodes. We report the average performance over three runs for each task, using three different random seeds.

\paragraph{SimplerEnv~\cite{li24simpler}} simulated benchmark was aimed at minimizing the gap between reality and simulation by optimizing the PD parameters with simulated annealing to minimize the gap between real and simulated end-effector trajectories. The evaluation is focused on four built-in tasks: pick coke can, move object near object, open drawer, and close drawer. Further, SimplerEnv allows two types of evaluation: (i) \emph{visual matching}, where the success of a task is determined by superimposing the real-life images on the simulation background and (ii) \emph{variant aggregation}, where the success rate of a task is averaged over the many variants of evaluation environment that differ in lighting, background, textures, distractor objects, etc. The full evaluation suite contains more than 1,000 episodes across these built-in tasks. We report the average performance over two runs for each task.

\paragraph{Cross-Embodiment Evaluation.}
To evaluate our model in the real-world, we assessed it on a Galaxea A1 robotic arm. This embodiment was deliberately chosen due to its absence from the large-scale pretraining dataset~\cite{open_x_embodiment_rt_x_2023}.
To adapt to this embodiment, we first collected 1,000 episodes of Pick-and-Place tasks via teleoperation. During data collection, we randomize the location of the objects in each episodes to enforce spatial generalization. We collected nine unique tasks, such as, \textit{``Put apple on the plate'', ``Put mango in the basket''}, and \textit{``Move the banana next to the plate''}. This set of tasks was designed to cover a variety of common objects.

To validate the performance of our models, we designed nine tasks to perform evaluation. Each task is repeated for $10$ trials, adhering to \citet{kim2024openvla}. To ensure a rigorous and fair comparison, these trials used 10 different fixed starting positions, which were kept consistent across all baselines. We divided our nine evaluation tasks into three categories, with three tasks per category. The first category consists of ``seen'' tasks, which are tasks that were also included in our fine-tuning dataset. This aims to validate the performance of our model and other baselines by cross-embodiment transfer. 

The second category, ``Unseen Object with Seen Distractor'', features tasks like ``Put X in plate''. Here, the target `X' is an unseen object absent from the fine-tuning dataset, while a familiar ``seen'' object `Y' present in the fine-tuning dataset is simultaneously placed in the environment as a distractor. This setup aims to evaluate the models' ability to generalize and their instruction following capabilities. 

The final category, ``Unseen Instruction with Seen Distractor'' features tasks like ``Move X to Z''. These tasks consist of simple instructions absent from the fine-tuning set and require the model to manipulate a novel object `X' (unseen in fine-tuning) and place it relative to a ``seen'' destination object `Z'. Crucially, a separate ``seen'' object `Y' is also present in the scene as a distractor. This setup aims to evaluate the models' ability to generalize to out-of-distribution instructions and their robustness to the presence of distractors.



\paragraph{Metric.}
In the LIBERO and SimplerEnv simulations, if the robot successfully completes the task specified by the prompt, then the trial is counted as a success, receiving a score of $1$; otherwise, a score of $0$ is assigned:
\begin{flalign*}
    & \text{\% success rate}_\tau \coloneq \\
    & (100 \mathbb{E}_{\text{trial} \sim \{1, 2,\cdots, 10\}} \mathbf{1}[\text{task $\tau$ is successfully completed}]) \%.
\end{flalign*}

For the Cross-Embodiment Evaluation, we report both success rate and partial success rate. The partial success metric is crucial for this real-world setting, as it allows us to differentiate between models that fail completely and those that make significant progress, thereby providing a more comprehensive breakdown of performance and failure modes.

%% file: related-works.tex
\section{Related Works}

\label{sec:related}

\paragraph{Vision–Language–Action Models.}
Large-scale vision–language–action (VLA) models learn general robot policies by training transformer policies on diverse demonstration datasets.
RT-1~\cite{brohan2023rt1} and the RT-X family trained on the Open X-Embodiment dataset~\cite{openxembodiment2023rtx} demonstrated that scaling real-world robot data and model capacity yields strong generalization across tasks and embodiments.
Subsequent open VLA models follow this recipe while incorporating stronger vision–language backbones, including OpenVLA~\cite{kim2024openvla}, SpatialVLA~\cite{qu2025spatialvla}, TraceVLA~\cite{zheng2024tracevla}, NORA~\cite{hung2025nora}, Emma-X~\cite{sun2024emmax}, EO-1\cite{qu2025eo1interleavedvisiontextactionpretraining}, and MolmoAct~\cite{lee2025molmoact}.
These approaches primarily rely on supervised imitation learning on large cross-embodiment datasets, sometimes with additional embodiment-specific fine-tuning, but they do not study reward-based post-training of VLA policies.

Orthogonally, flow-matching-based action models such as $\pi_0$~\cite{black2024pi0} and $\pi_{0.5}$~\cite{physicalintelligence2025pi05} attach a continuous-time flow-matching action head to a pre-trained vision--language backbone to generate smooth, real-time continuous action trajectories. In parallel, discretized action tokenization methods such as FAST~\cite{pertsch2025fast} focus on compressing continuous action sequences into short sequences of discrete tokens for efficient autoregressive decoding, and can be combined with $\pi_0$ to obtain the $\pi_0$-FAST variant. Our NORA-1.5 architecture similarly augments a pre-trained VLA backbone~\cite{hung2025nora} with a flow-matching-based action expert; unlike prior work, we find that this coupling not only improves inference speed~\cite{black2024pi0,physicalintelligence2025pi05} but also yields consistent accuracy gains across simulated and real-world benchmarks.

\paragraph{World Models for Visual Robot Control.}
Self-supervised video and world models aim to predict future observations conditioned on current observations and actions, and have been used for planning and model-based control~\cite{wu2024ivideogpt,guo2025ctrlworld,zheng2025flare}.
V-JEPA2~\cite{assran2025vjepa2} learns a latent video prediction objective that can be extended to action-conditioned dynamics, while DINO-WM~\cite{zhou2025dinowm} performs planning in the latent space of a pre-trained visual encoder.
These methods typically use the world model online for planning or trajectory optimization.
In contrast, we repurpose an action-conditioned V-JEPA2 variant as a reward model that scores full action sequences.
This enables scalable synthetic preference generation for VLA post-training without task-specific reward engineering or high-fidelity robot simulators.
Sim2real evaluation frameworks such as SimplerEnv~\cite{li2024simplerenv} focus on accurately matching real robot trajectories in simulation to provide reliable evaluation for manipulation policies; we adopt such benchmarks to assess the gains from our world-model-based rewards.

\paragraph{Preference-based Post-Training and Reward Design.}
Preference optimization has become a standard tool for aligning large language models with human intent.
Direct Preference Optimization (DPO)~\cite{rafailov2023dpo} optimizes a policy directly from pairwise preferences, and Group Relative Preference Optimization (GRPO)~\cite{shao2024deepseekmath} extends this idea to group-wise comparisons.
While VLA models are usually trained purely with supervised imitation learning, recent work such as RoboMonkey~\cite{kwok2025robomonkey} explores synthetic reward signals for test-time sampling and verification of robot actions, without updating the underlying policy.
Our work brings preference-based post-training to the VLA setting by constructing synthetic preferences from two complementary reward signals: a world-model-based goal reward and a distance-to-expert-action heuristic.
We show that combining these rewards with DPO yields consistent performance improvements over purely supervised training on LIBERO~\cite{liu2023libero} and SimplerEnv~\cite{li2024simplerenv} benchmarks.